\documentclass[11pt]{article}
\usepackage{amsmath,amsfonts}
\usepackage{algorithmic}
\usepackage{algorithm}
\usepackage{array}
\usepackage[caption=false,font=normalsize,labelfont=sf,textfont=sf]{subfig}
\usepackage{textcomp}
\usepackage{stfloats}
\usepackage{url}
\usepackage{verbatim}
\usepackage{graphicx}
\usepackage{cite}
\usepackage{supertabular}			
\usepackage{soul}	

\graphicspath{{./figures/}}

\usepackage{multirow}		
\usepackage{multicol}

\usepackage{colortbl}
\definecolor{sandyBrown}{rgb}{0.95,0.64,0.38}
\definecolor{lightSkyBlue}{rgb}{0.53,0.81,0.98}
\definecolor{lightGreen}{rgb}{0.56,0.93,0.56}

\usepackage{comment}

\DeclareMathOperator{\sinc}{sinc}

\renewcommand{\frame}[1]{{\cal F}_{#1}}

\newcounter{tableCounter}

\newcommand{\bt}[1]{\textcolor{magenta}{#1}}

\begin{document}

\title{Safe Path following for Middle Ear Surgery}

\author{Bassem Dahroug$^{1}$, Brahim Tamadazte$^{2}$, and Nicolas Andreff$^{1}$ \\
\\
$^{1}$Bassem Dahroug and Nicolas Andreff are with FEMTO-ST, AS2M, \\ Univ. Bourgogne Franche-Comt\'e, \\CNRS/ENSMM, 25000 Besan\c con, France\\
$^{2}$Brahim Tamadazte is with Sorbonne Université,\\ CNRS UMR 7222, INSERM U1150, ISIR, F-75005, Paris, France.\\
\texttt{brahim.tamadazte@cnrs.fr}
}


\maketitle
\begin{abstract}
This article formulates a generic representation of a path-following controller operating under contained motion, which was developed in the context of surgical robotics. It reports two types of constrained motion: i) Bilateral Constrained Motion, also called Remote Center Motion (RCM), and ii) Unilaterally Constrained Motion (UCM). In the first case, the incision hole has almost the same diameter as the robotic tool, while in the second state, the diameter of the incision orifice is larger than the tool~diameter. The second case offers more space where the surgical instrument moves freely without constraints before touching the incision wall.

The proposed method aims to combine two tasks that must operate hierarchically: i) respect the RCM or UCM constraints formulated by equality or inequality, respectively, and ii) perform a surgical assignment, e.g., scanning or ablation expressed as a 3D path-following task.
The proposed methods and materials were successfully tested first on our simulator that mimics realistic conditions of middle ear surgery, then on an experimental platform. Different validation scenarios were carried out experimentally to assess quantitatively and qualitatively each developed approach. 
Although ultimate precision was not the goal of this work, our concept is validated with enough accuracy ($\leq 100 \mu m$) for the ear~surgery.

\textbf{keywords}: Medical Robotics, Constrained motion, Path Following, Visual Servoing.
\end{abstract}
%
\section{INTRODUCTION}
Surgical robots are gaining more popularity due to their advantages for both the patient and the physician~\cite{nageotte2006path, dahroug2018review, troccaz2019frontiers}.
It is particularly valid for so-called Minimally-Invasive Surgery (MIS) approaches. For instance, a laparoscopy or keyhole surgery~\cite{lee2018_rcmMIS} performs incision around $10mm$. It is tiny compared to the larger incisions needed in laparotomy (open surgery). Another situation where the surgical instruments could be inserted through a natural orifice (e.g., mouth, nasal clefts, urethra, anus) to reach the targeted organ. In both cases, the entry space (i.e., the incision hole or the natural orifice) restricts the surgical tool motion, consequently the surgeon's hands and the robot carrying the instrument~\cite{bowyer2014_surveyVF}. 

This article mainly discusses two types of constrained motion that result directly from MIS procedures: 
\begin{enumerate}
	\item \textit{Remote Center Motion} (RCM), also known as \emph{fulcrum effect}, implies the incision hole has almost the same diameter as that of the surgical tool~\cite{aksungur2015_surveyRCM};
	\item \textit{Unilaterally Center Motion} (UCM) implies the incision diameter size is bigger than that of the tool, offering more freedom for the tool motion~\cite{dahroug2019_UCM}.
\end{enumerate}

The first type of motion was initially achieved by designing a particular robotic structure that imposes the constrained motion mechanically~\cite{kuo2009review, aksungur2015_surveyRCM, nisar2020}. The RCM dictates that the center-line of the surgical tool is always coincident with the center point of the incision orifice (trocar point). Consequently, the linear movement of the tool is prohibited along two axes. 
The main advantage of RCM mechanisms is to reduce the risk of damaging the trocar wall because their kinematic structures ensure the pivoting motion. Their modest controller is also easy to implement. However, this kind of mechanism is restricted to a unique configuration and cannot provide enough flexibility for shifting the location of the trocar~point. 

An alternative solution proposes a software RCM for overcoming the previous problem by guiding a general-purpose robot with the advantage of being flexible enough for achieving complex tasks~\cite{boctor2004virtualRCM}. This solution is convenient for diverse medical applications (e.g., laparoscopic~\cite{osa2010_rcmVS} and eye~\cite{nasseri2014_VFOpthalmic} surgeries). However, we claim that the RCM approach is not the best choice for other surgery types (e.g., ear, nose, mouth, knee arthroscopy). In latter cases, the orifice diameter is generally bigger than the tool diameter. Consequently, the RCM controller imposes too strong limitations on the tool motion. Indeed, the RCM is a mathematical equality constraint (i.e., the distance between the tool body and the center point of the incision orifice must be equal to zero). As such, RCM motion can be named as a bilaterally constrained motion. On the contrary, UCM is a weaker restriction since the unilateral constraints are inequality equations (i.e., the latter distance could be greater or less than zero)~\cite{kanoun2011_inequalityTask}.

In the literature, the term forbidden-region virtual fixtures~\cite{abbott2007_hapticVF} are used for collaboration tasks where the user can either manipulate a robotic device~\cite{becker2013_visionVF} or telemanipulate a master device~\cite{selvaggio2018_passiveVF}. 
These fixtures could be defined as geometric forms~\cite{zheng2018_forbiddenVF, dahroug2019_UCM} or vector field~\cite{marinho2020virtual} around the tool. Then a kinematic control~\cite{dahroug2019_UCM} or dynamic one~\cite{vitrani2016VF, zheng2018_forbiddenVF, marinho2020virtual} is applied to guide the robot during the desired task.

The theoretical contribution of this article lies in the improvement of the generic formulation of constrained motion. It has the objective to achieve a velocity controller that can maintain the RCM or UCM depending on the configuration of the surgical procedure.
Besides that, it reveals a new path-following controller integrated with a task-hierarchy controller for imposing a priority between the RCM/UCM and the path-following tasks. 

Nevertheless, the technical contribution lies in the assessment of such approaches. Therefore, we developed a simulator including surgical tools and a numerical twin mimicking the middle ear cavity. 
Based on the auspicious evaluation, we also carried out a pre-clinical setup that takes up the diverse components of the simulator to assess the proposed methods experimentally. Various scenarios are also implemented to accomplish these evaluations. The obtained performances in terms of behavior and accuracy are promising.

The remainder of the article is organized as follows. Section~\ref{sec: motivation} presents the clinical needs and challenges. The methodology followed to design the proposed controllers will be discussed in Section~\ref{sec: methodology}. After that, Section~\ref{sec: validation} focuses on both the numerical and experimental validations of the proposed approaches. Ultimately, Section~\ref{sec: conclusion} presents the conclusion and~perspectives. 

\section{MEDICAL MOTIVATIONS}	\label{sec: motivation}
%
\subsection{Treated Disease}
The work discussed in this article represents a part of a long-term project. It deals with the development of a robotic system that is dedicated to cholesteatoma surgery. The system will aim to achieve an MIS within the middle ear cavity by passing through the external ear canal or an incision orifice made on the mastoid portion. 

\begin{figure}[!h]
	\centering
	\includegraphics[width=.8\columnwidth]{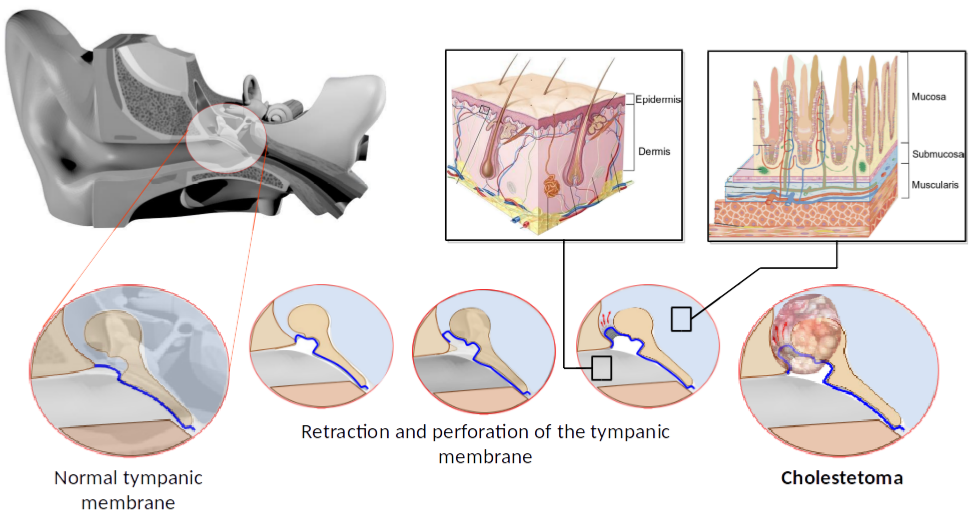}
	\caption{Evolution of cholesteatoma disease within the middle ear, which is located behind the tympanic membrane.}
	\label{fig: cholesteatomaEvol}
\end{figure}

Cholesteatoma is a frequent disease that invades the middle ear. It infects the middle ear by introducing abnormal skin (lesional tissue) in the middle ear-cavity. The most common explanation~\cite{olszewska2004_etiopathogenesis} is due to the immigration of the epidermal cells, which are the cells type in the external ear canal, and cover up the mucosa of the middle ear cavity, as shown in Fig.\ref{fig: cholesteatomaEvol}. These cells gradually proliferate within the temporal bone and destroy the adjacent bony structures. 
%
%
%
%

The evolution of cholesteatoma is life-threatening in the long run. The complications can be classified as follows~\cite{alper2004advanced}: i) destruction of the ossicular chain, ii) facial paralysis, iii) labyrinthitis, iv) extracranial complications, and v) intracranial complications. It can notice the irreversible effects that cholesteatoma can cause in a patient. Despite that, there is no drug therapy for the treatment. The only solution is surgical~intervention.

\subsection{Current Surgical Procedure}
As claimed above, the only treatment for cholesteatoma is a surgical procedure. It aims to eradicate all cholesteatoma tissue and reconstruct the anatomy of the middle ear~\cite{hildmann2006middle}.

\begin{figure}[!h]
	\centering
		\includegraphics[width=.6\columnwidth]{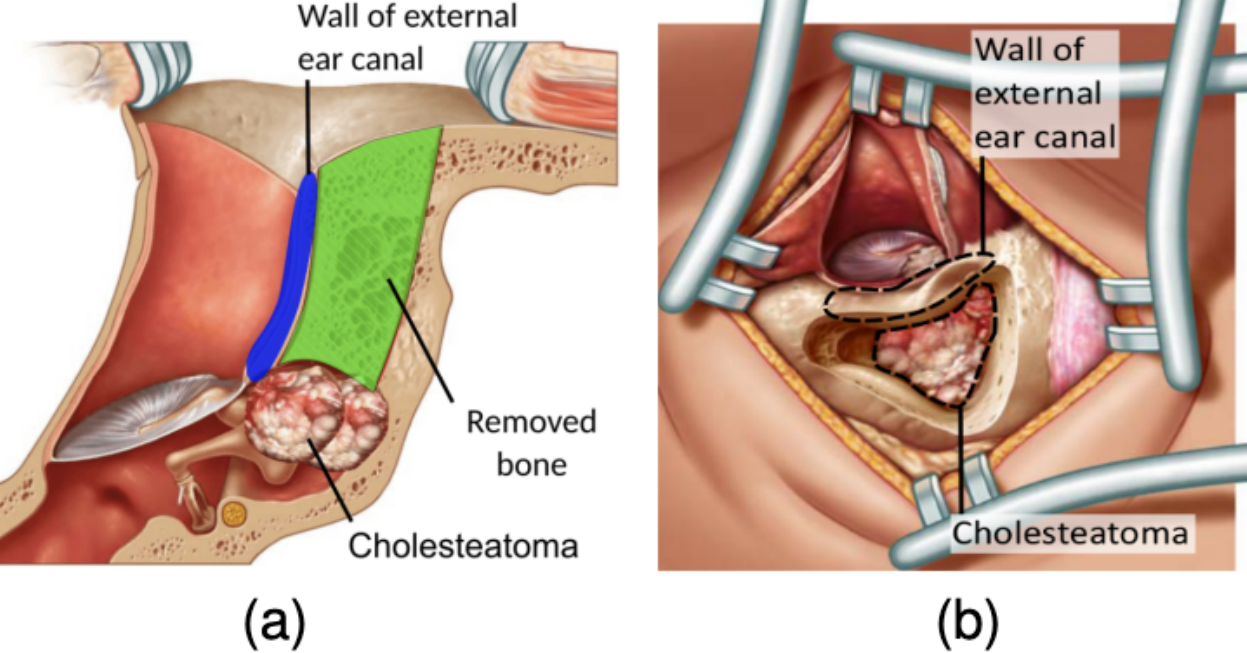}
	\caption{Mastoidectomy procedure with canal-wall-up indicates that the external ear canal is preserved. (a) side view of the mastoidectomy tunnel and (b) top view of the mastoidectomy tunnel.}
	\label{fig: mastoidectomyCWU}
\end{figure}
For reaching the middle-ear cavity, the surgeon often drills the temporal bone behind the auricular, as shown in Fig.~\ref{fig: mastoidectomyCWU}. This surgical procedure is called \textit{mastoidectomy} where the surgeon maintains the wall of the external ear canal. This technique creates an incision that forms a triangular (around $40\times40\times30mm$) with a depth of about $30mm$. The latter procedure can also become more invasive by sacrificing the posterior portion of the external ear canal (i.e., \textit{canal-wall-down}).  
Furthermore, even if the surgical orifice is relatively large, the surgical procedure remains complex and requires high expertise and dexterity from the surgeon. Also, even with an experienced clinician in the cholesteatoma case, the clinical outcomes remain unsatisfactory in terms of effectiveness. Indeed, there is a high risk that the cholesteatoma could regrow a few months after the surgical intervention. It occurs due to residual cholesteatoma cells. Consequently, $10$ to $40\%$ of patients perform more than one surgery to get definitively over this disease\cite{aquino2011epidemiology}.

\begin{figure}[!h]
	\centering
	\includegraphics[width=.7\columnwidth]{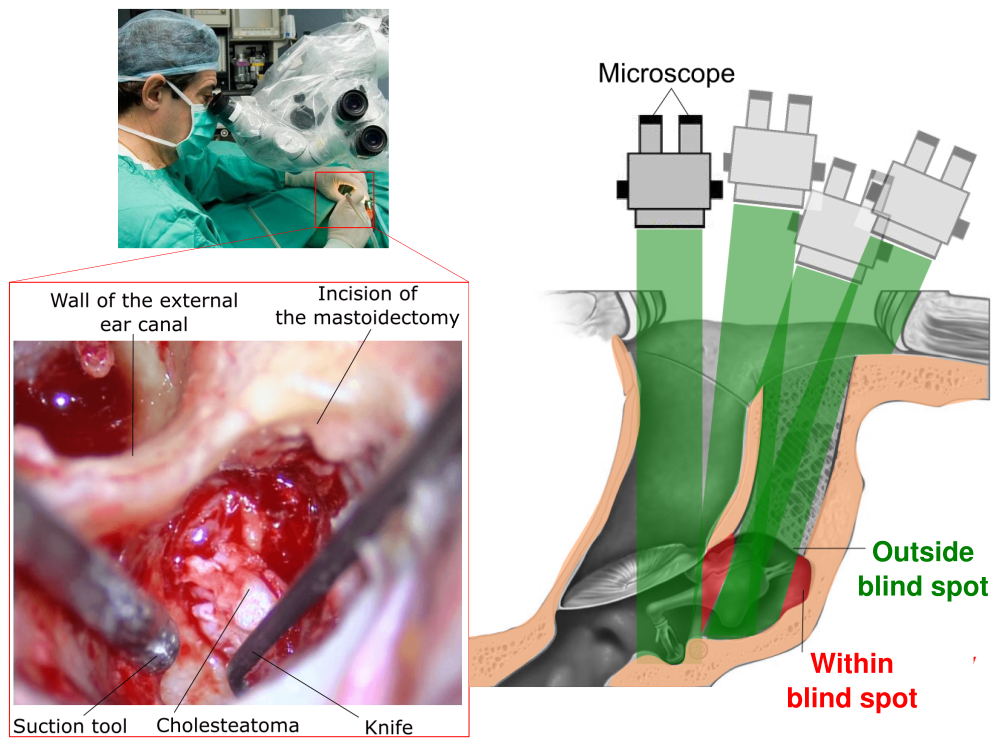}
	\caption{Conceptual scheme to demonstrate the "blind spot" during the cholesteatoma surgery.}
	\label{fig: mastoidectomyDifficulties}%
\end{figure}	
Due to the complexity of the temporal bone cavity, the surgeon mainly faces numerous difficulties during the procedure (Fig.\ref{fig: mastoidectomyDifficulties}): i) lack of ergonomy of the tools; ii) limited field of view of the oto-microscope (the surgeon cannot visualize the lateral regions hidden (\textit{blind spots}) in the middle ear cavity) and iii) access with the conventional rigid instruments requires considerable expertise to handle.

Therefore, it is increasingly important to overcome the previous problems and evolve this procedure towards less invasive. It implies reducing the incision orifice size, improving the cholesteatoma ablation efficiency, and avoiding the current high surgical recurrence rate for this kind of surgery.

\section{METHODOLOGY}\label{sec: methodology}
This section begins by presenting a brief summary of the new surgical protocol associated with the robotic system. After that, it discusses the hierarchical controller for managing simultaneously the various tasks. It then explains separately the path-following, the RCM, and the UCM controllers.

\subsection{New Surgical Protocol}
In collaboration with surgeons experts in middle ear surgery, especially cholesteatoma treatment, we have attempted to set up a new and more efficient surgical protocol reported in~\cite{dahroug2018review}.
Firstly, the idea is to make cholesteatoma surgery less invasive compared to the traditional one. Thus, a macro-micro robotic system should pass through a millimetric incision made behind the ear (in the mastoid portion) to access the middle ear cavity~\cite{so2020micro}.  
Secondly, cholesteatoma surgery needs to be more efficient by eliminating the residual cases. This second objective can be accomplished by removing a large part of the cholesteatoma tissue using rigid miniature mechanical resection tools. After that, a bendable actuated tool~\cite{gafford2020eyes, swarup2021design} could be used to guide a laser fiber. This fiber carbonizes the residual cholesteatoma (resulting from the mechanical resection phase)~\cite{kimberley2020}.

Both mechanical resection and laser ablation should be performable either in automatic or semi-automatic mode. While the mechanical resection does not require high accuracy, the laser ablation requires higher precision since the residual cholesteatoma cells can be a few tens of micrometers in size. Therefore, the contributions of robotics and vision-based control are essential to fundamental this kind of task. In this work, we investigated the use of path-following control schemes under constrained motion (due to the incision orifice) to carry out the notions requested by the cholesteatoma removal (i.e., mechanical resection and laser ablation).

\subsection{Task Hierarchical Controller}	\label{subSec: taskHierarchy}
A surgical procedure can be considered as a set of sequential or overlapping sub-tasks. The hierarchical methods ensure the execution of several tasks simultaneously. Consequently, the required tasks do not enter into conflict~\cite{siciliano1990kinematic, escande2014hierarchical}. In the case of cholesteatoma surgery, various sub-tasks can be involved during the procedure, such as constraint enforcement (RCM or UCM) and ablation tools for the pathological tissues. Therefore, these sub-tasks must be carried out according to a defined hierarchical scheme.

To express a controller that manages simultaneous sub-tasks, let us start by assuming that a generic sub-task ($\mathbf{\dot{e}}_i \in \mathbb{R}^{m_i}$) given by
\begin{equation}		\label{eq: th_task}
\mathbf{\dot{e}}_i = \mathbf{L}_i ~^e\mathbf{\underline{v}}_e, ~~~\text{where i=1,2,...,j} 
\end{equation}
where $^e\mathbf{\underline{v}}_e \in se(3)$ is the end-effector twist velocity to be computed in the end-effector frame $\frame{e}$,
and $\mathbf{L}_i \in \mathbb{R}^{m_i \times n}$ is the interaction matrix which relates the vector $^e\mathbf{\underline{v}}_e$ to the error~$\mathbf{\dot{e}}_i$.

The inverse solution of the previous equation is not guaranteed since the interaction matrix $\mathbf{L}_i$ could be non-square, and the matrix rank is locally deficient. Thanks to the least-square method, an approximate solution can be found by minimizing $\Vert \mathbf{\dot{e}}_i - \mathbf{L}_i ~^e\mathbf{\underline{v}}_e \Vert$ over $^e\mathbf{\underline{v}}_e$, and using numerical procedures (such as QR or SVD).
The formal result of it can be simply written as $^e\mathbf{\underline{v}}_e = \mathbf{L}_i^\dagger \mathbf{\dot{e}}_i$, where $\mathbf{L}_i^\dagger$ is the pseudo-inverse of $\mathbf{L}_i$. If $\mathbf{L}_i$ does not have full rank then it has at least one singular vector $\mathbf{z}_1$, located in its null-space ($\mathbf{L}_i \mathbf{z}_1 = \mathbf{0}$). The vector $\mathbf{z}_1$ is also described as the null space of $\mathbf{e}_i$, because any twist vector parallel to $\mathbf{z}_1$ will leave $\mathbf{e}_i$ unchanged.
Therefore, the projection gradient general form~\cite{nakamura1987taskPriority} is given by
\begin{equation}		\label{eq: th_generalForm0}
^e\mathbf{\underline{v}}_e = \mathbf{L}_1^\dagger \mathbf{\dot{e}}_1 + (\mathbf{I} - \mathbf{L}_1^\dagger \mathbf{L}_1) \mathbf{z}_1 \quad  
\end{equation}

In order to define $\mathbf{z}_1$, let us first consider a secondary sub-task $\mathbf{\dot{e}}_2 = \mathbf{L}_2 ~^e\mathbf{\underline{v}}_e$. Since the control vector must include the first sub-task, equation (\ref{eq: th_generalForm0}) is injected in the latter expression, resulting in
\begin{equation}
\begin{split}
\mathbf{\dot{e}}_2 &= \mathbf{L}_2 \left( \mathbf{L}_1^\dagger \mathbf{\dot{e}}_1 + (\mathbf{I} - \mathbf{L}_1^\dagger \mathbf{L}_1) \mathbf{z}_1 \right) \\
&= \mathbf{L}_2 \mathbf{L}_1^\dagger \mathbf{\dot{e}}_1 + \underset{\mathbf{\tilde{L}}_2}{\underbrace{\mathbf{L}_2 (\mathbf{I} - \mathbf{L}_1^\dagger \mathbf{L}_1)}} \mathbf{z}_1 \quad 
\end{split}
\end{equation}
From the previous equation, the vector $\mathbf{z}_1$ is deduced as
\begin{equation}		\label{eq: th_optimalVector1}
\mathbf{z}_1  = \mathbf{\tilde{L}}_2^\dagger (\mathbf{\dot{e}}_2 - \mathbf{L}_2 \mathbf{L}_1^\dagger \mathbf{\dot{e}}_1) + (\mathbf{I} - \mathbf{\tilde{L}}_2^\dagger \mathbf{\tilde{L}}_2) \mathbf{z}_2 
\end{equation}
with another criteria vector $\mathbf{z}_2$ which is projected in the null-space of the secondary sub-task.
By introducing (\ref{eq: th_optimalVector1}) in (\ref{eq: th_generalForm0}), a recursive form of the projection gradient is obtained~as
\begin{equation}		\label{eq: th_generalForm1}
\small
\begin{split}
^e\mathbf{\underline{v}}_e &= \mathbf{L}_1^\dagger \mathbf{\dot{e}}_1 + (\mathbf{I} - \mathbf{L}_1^\dagger \mathbf{L}_1) \left( \mathbf{\tilde{L}}_2^\dagger (\mathbf{\dot{e}}_2 - \mathbf{L}_2 \mathbf{L}_1^\dagger \mathbf{\dot{e}}_1) + (\mathbf{I} - \mathbf{\tilde{L}}_2^\dagger \mathbf{\tilde{L}}_2) \mathbf{z}_2 \right) \\
							&= \mathbf{L}_1^\dagger \mathbf{\dot{e}}_1 + (\mathbf{I} - \mathbf{L}_1^\dagger \mathbf{L}_1) \mathbf{\tilde{L}}_2^\dagger (\mathbf{\dot{e}}_2 - \mathbf{L}_2 \mathbf{L}_1^\dagger \mathbf{\dot{e}}_1) \\
							&~~~ + (\mathbf{I} - \mathbf{L}_1^\dagger \mathbf{L}_1) (\mathbf{I} - \mathbf{\tilde{L}}_2^\dagger \mathbf{\tilde{L}}_2) \mathbf{z}_2  \quad
\end{split}
\end{equation}
The right-hand side of the previous equation can further be simplified as \cite{maciejewski1985}
\begin{equation}		\label{eq: th_generalForm2}
^e\mathbf{\underline{v}}_e = \mathbf{L}_1^\dagger \mathbf{\dot{e}}_1 +  \mathbf{\tilde{L}}_2^\dagger (\mathbf{\dot{e}}_2 - \mathbf{L}_2 \mathbf{L}_1^\dagger \mathbf{\dot{e}}_1) \quad .
\end{equation}

The latter equation finds a solution to satisfy both sub-tasks $\mathbf{\dot{e}}_1$ and $\mathbf{\dot{e}}_2$. It also ensures a form of hierarchy/priority between them. 
The analytical expression of each sub-task with its $\mathbf{L}_i$ is presented in the coming sections.

\subsection{6D Approach Controller}     \label{subSec: approachTask}
This section is dedicated to mathematically describing how to control the tool-tip for regulating its position and orientation with respect to a reference frame, e.g., the orifice frame $\frame{r}$. This task is applied when the tool locates outside the incision orifice, and its pose must be adjusted with respect to the orifice before it starts another task inside the orifice.

To do this, a traditional 3D position-based visual servo~\cite{chaumette2006} is applied. The feature vector $\mathbf{s}~=~(^r\mathbf{t}_t, \theta ~^r\mathbf{u}_t)$ is defined as the pose vector which describes the tool-tip frame $\frame{t}$ with respect to the orifice frame $\frame{r}$. This vector gathers the translation $\mathbf{t}$ of the tool-tip and its rotation $\theta \mathbf{u}$ in form of angle/axis parameterization. The desired feature vector $\mathbf{s}^*~=~(\mathbf{0}, \mathbf{0})$ is set to a zero vector since it is required to make coincident the frame $\frame{t}$ with $\frame{r}$. Thus, the approach task error $\mathbf{e}_{app}$ is deduced as the difference between the current features vector and the desired one, i.e.,
\begin{equation} \label{eq: approachError}
\mathbf{e}_{app} = \mathbf{s} - \mathbf{s}^* \quad
\end{equation}

The time variation of the latter error is related to the spatial velocity of the tool-tip $^t\mathbf{\underline{v}}_t$ by the interaction matrix $\mathbf{L}_{3D} \in \mathbb{R}^{6\times6}$ as
\begin{equation}
\mathbf{\dot{e}}_{app} = \mathbf{L}_{3D} ~^t\mathbf{\underline{v}}_t
\end{equation}
where $^t\mathbf{\underline{v}}_t = (^t\mathbf{v}_t, ^t\mathbf{\omega})$ gathers the instantaneous linear and angular velocities of the tool-tip.
Since the desired feature vector equals to $\mathbf{0}_{6\times1}$, then the interaction matrix $\mathbf{L}_{3D}$ is determined by
\begin{equation}
\mathbf{L}_{3D} = \left[ \begin{array}{cc}
-\mathbf{I}_{3\times3} & \mathbf{0}_{3\times3} \\
\mathbf{0}_{3\times3} & \mathbf{L}_{\theta \mathbf{u}}
\end{array} \right]
\end{equation}
where $\mathbf{I}_{3\times3}$ is a $3\times3$ identity matrix, $\mathbf{0}_{3\times3}$ is a $3\times3$ zero matrix, and $\mathbf{L}_{\theta \mathbf{u}}$ is given by~\cite{malis1999}
\begin{equation}
\mathbf{L}_{\theta \mathbf{u}} = \mathbf{I}_{3\times3} - \frac{\theta}{2} \left[ \mathbf{u} \right]_{\times} + \left( 1 - \frac{\sinc \theta}{\sinc^2 \frac{\theta}{2}} \right) \left[ \mathbf{u} \right]^2_{\times}
\end{equation}
in which $\sinc x$ is the sinus cardinal.

Finally, the spatial velocity $^t\mathbf{\underline{v}}_t$ is determined for ensuring an exponential decoupled reduction of the error (i.e., $\mathbf{\dot{e}} = - \lambda \mathbf{e}$)~as
\begin{equation}	\label{eq: controlApp}
^t\mathbf{\underline{v}}_t = - \gamma \mathbf{L}_{3D}^{-1} \mathbf{e}_{app}
\end{equation}
where $\gamma$ is a gain coefficient, and $\mathbf{L}_{3D}^{-1}$ is the inverse of the interaction matrix since it is square and has a closed-form inverse~\cite{malis1999}. 

The command velocity of the robot end-effector $^e\mathbf{\underline{v}}_e = ~^e\mathbf{V}_t ~^t\mathbf{\underline{v}}_t$ is deduced by the following twist~matrix
\begin{equation}
^e\mathbf{V}_t = \left[ \begin{array}{cc}
~^e\mathbf{R}_{t} & \left[ ^e\mathbf{t}_t \right]_{\times} ~^e\mathbf{R}_{t}\\
\mathbf{0}_{3\times3} & ~^e\mathbf{R}_{t}
\end{array} \right]
\end{equation}
since the tool body is rigid and the transformation between the end-effector frame $\frame{e}$ and the tool-tip frame $\frame{t}$ is fixed. Finally, the controller stability was demonstrated in~\cite{malis1999} to be globally exponentially stable.

\subsection{3D Path-Following Controller}\label{subSec: pf}
This section will focus on a generic modelling of a 3D path-following scheme. The advantage of using such as controller is the separation between i) the geometric curve (desired path $\mathcal{S}_p$) which is planned by the surgeon based on pre-operative images, and ii) the advance speed ($v_{tis}$) of the tool-tip along the desired path which is controlled by the surgeon during the operation. 
In this manner, the collaboration surgeon/robot ensures that the robot guides the tool along the path while the surgeon controls the robot progression without planning the robot velocity direction.

\begin{figure}[!h]
	\centering
	\includegraphics[width=0.6\columnwidth]{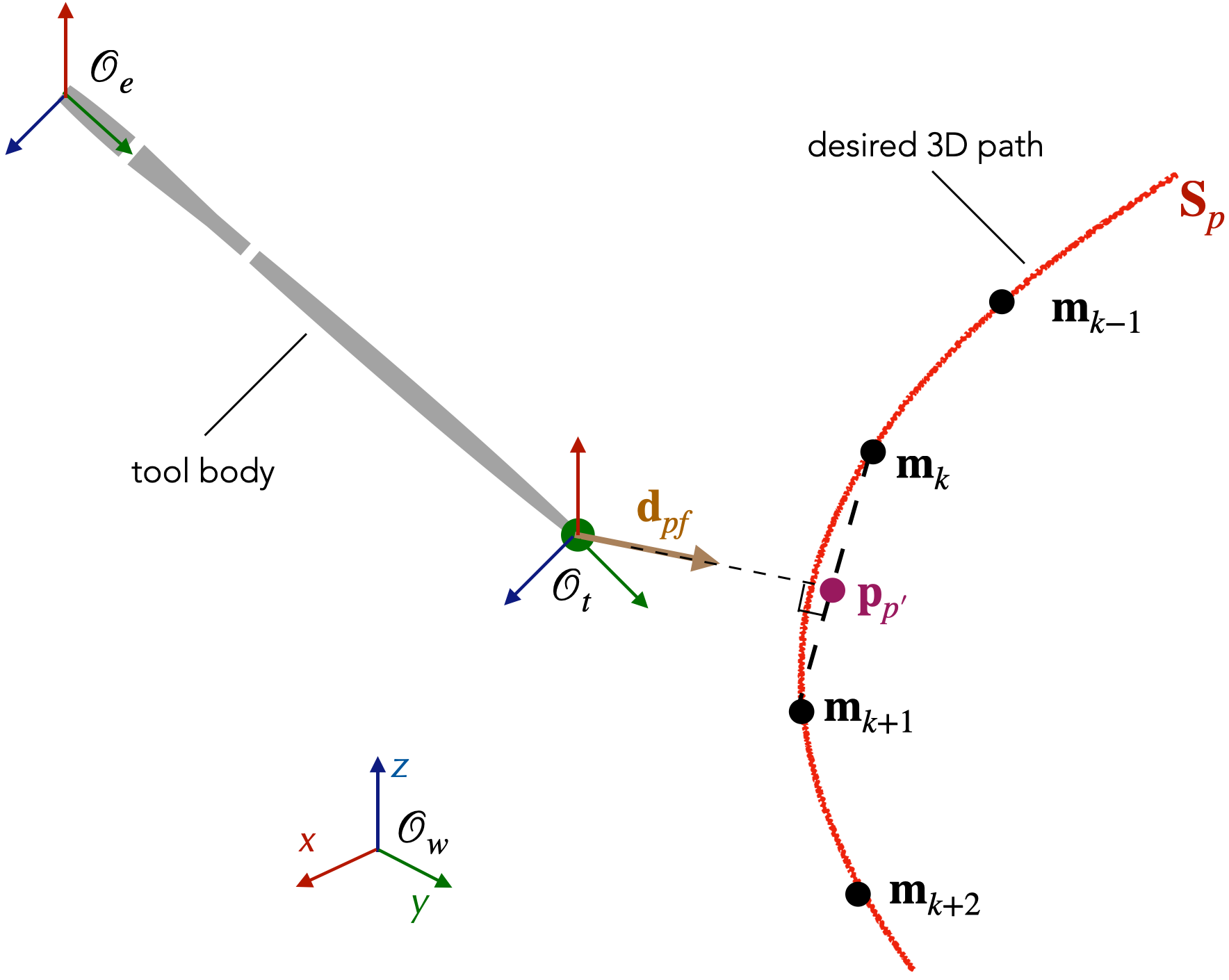}
	\caption{Orthogonal projection of the tool-tip onto a geometric curve. } 
	\label{fig: path_proj}
\end{figure}	
Fig.~\ref{fig: path_proj} depicts the surgical instrument and its reference frames with respect to the desired path $\mathcal{S}_p$. By projecting the tool-tip $\mathcal{O}_{t}$ onto the reference path, the resultant orthogonal distance $\mathbf{d}_{pf}$ is considered as the error (i.e., lateral deviation) which must be controlled to zero. Therefore, the 3D vector distance between the tool-tip $\mathcal{O}_{t}$ and the projection point $\mathbf{p}_{p^\prime}$ calculated as
\begin{equation}  \label{eq: d_pf}
\mathbf{d}_{pf} = \mathcal{O}_{t} - \mathbf{p}_{p^\prime}.
\end{equation}

In order to express the command velocity, the time-derivative of (\ref{eq: d_pf}) provides the tool-tip velocity $\mathbf{v}_{t}$ as discussed in~\cite{dahroug2017_vsPFRCM}
\begin{equation}
\label{eq: dot_d_pf(LV_t)} 
\mathbf{\dot{d}}_{pf} = \left( \mathbf{I}_{3\times3} - \frac{\mathbf{k}_{p} \mathbf{k}_{p}^\top}{1 - \mathbf{d}_{pf}^\top \Big(\mathbf{C}_p(s_p) \times \mathbf{k}_{p}\Big)} \right) \mathbf{v}_t 
\end{equation}
where $\mathbf{C}_p(s_p)$ is the path curvature in function of the path curve length, $\mathbf{k}_{p}$ is the unit-vector of the instantaneous tangential vector (Fig.~\ref{fig: path_proj}).

\begin{figure}[!h]
	\centering
	\includegraphics[width=0.6\columnwidth]{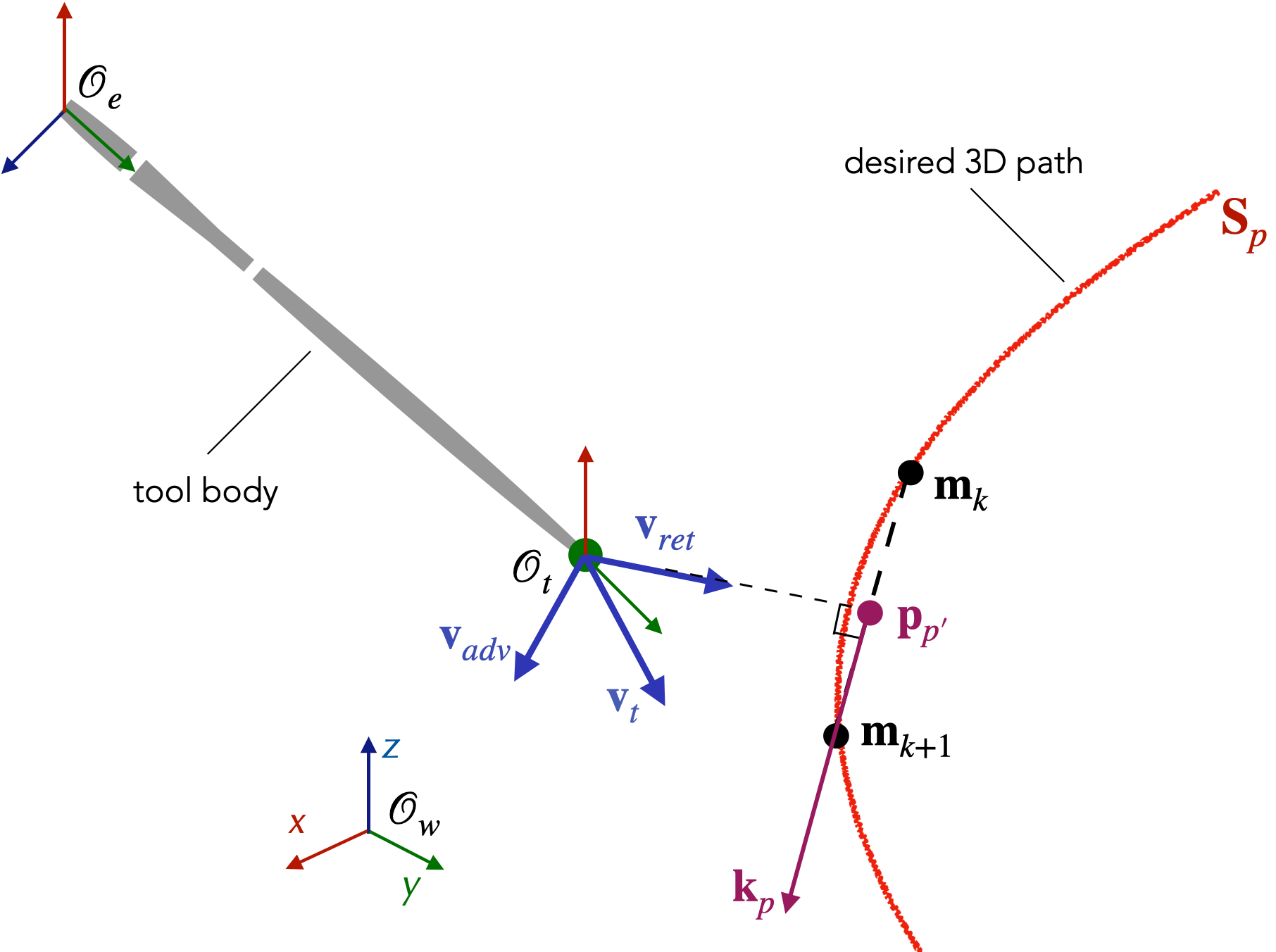}
	\caption{Representation of the different velocities involved in the path-following controller. }
	\label{fig: path_velocity}
\end{figure}

At this stage, it requires to choose the adequate velocity of the tool-tip $\mathbf{v}_t$ in the latter equation to ensure that the lateral error $\mathbf{d}_{pf}$ is regulated to zero while progressing along the path. An intuitive solution consists of decomposing the control velocity into two orthogonal components (Fig.~\ref{fig: path_velocity}): i) the advance velocity ($\mathbf{v}_{adv}$) along the path, and ii) the return velocity ($\mathbf{v}_{ret}$) for regulating the tool deviation from the reference path. The previous concept is formulated as follows:
\begin{equation} \label{eq: LV_t}
\mathbf{v}_t = \underset{\mathbf{v}_{adv}}{\underbrace{\alpha \mathbf{k}_p}} + \underset{\mathbf{v}_{ret}}{\underbrace{\beta \mathbf{d}_{pf}}} \quad .
\end{equation}

The tuning coefficients of the controller $\alpha$ and $\beta$ allow adjusting the priority between the advance and return velocities, respectively. Besides that, the controller stability demonstrated in~\cite{dahroug2017_vsPFRCM} shows that $\alpha$ should be a positive scalar while $\beta$ must be a negative scalar to ensure the system stability.

The choice of these gain factors can be imposed by a function of a constant velocity $v_{tis} > 0$ that depends on the interaction between the surgical tool and the lesional tissue. This velocity could be tuned easily by the surgeon before or during the intervention. 
Therefore, (\ref{eq: LV_t}) yields
\begin{equation} \label{eq: choice__LV_t}
\underset{= \Vert \mathbf{v}_{t} \Vert^2}{\underbrace{v_{tis}^2}} = \alpha^2 \underset{=1}{\underbrace{\Vert \mathbf{k}_p \Vert^2}} + \underset{= \Vert \mathbf{v}_{ret} \Vert^2}{\underbrace{\beta^2 \Vert \mathbf{d}_{pf}\Vert^2}} \quad .
\end{equation}

The gain factor $\alpha$ is thus determined as
\begin{equation} \label{eq: choice__alpha}
\alpha =  \left\lbrace 
\begin{array}{ll} 
\sqrt{v_{tis}^2 - \Vert \mathbf{v}_{ret} \Vert^2} & \Vert \mathbf{v}_{ret} \Vert^2 < v_{tis}^2 \\ 
0 & \Vert \mathbf{v}_{ret} \Vert^2 > v_{tis}^2 
\end{array} \right. \quad .
\end{equation}
If the tool is not far from the reference path, the first condition in (\ref{eq: choice__alpha}) is selected. Otherwise, the priority is returning the tool-tip to the reference path, and the advance velocity is null (i.e., second condition in (\ref{eq: choice__alpha})). 

The latter strategy proposed in \cite{dahroug2017_vsPFRCM} applies a constant value for the gain factor $\beta$. However, this section presents a new formulation of $\beta$ to make the controller sensitive to the path curvature. Thus, it is calculated by the following equation
\begin{equation} \label{eq: choice__beta} 
\beta = \beta^\prime \bigg ( 1 + sign \left( \mathbf{d}_{pf}^\top \left( \mathbf{C}_p(s_p) \times \mathbf{k}_p \right) \right) \left( 1 - e^{\gamma_{c} \Vert \mathbf{C}_p(s_p) \Vert} \right) \bigg)
\end{equation}
where $\beta^\prime$ is a negative gain for returning to path, $sign(\bullet)$ is a sign function to determine the direction along the reference path, and $\gamma_{c}$ is a negative gain for sensing the amount of path~curvature.

The ratio between the gain factors (i.e., $v_{tis}$ and $\beta^\prime$) forms an acceptable error band around the reference path. For instance, if $\beta^\prime$ is higher than $v_{tis}$, then the error band will be small. On the contrary, in the case where $v_{tis}$ is bigger than $\beta^\prime$, then the error band will be large since the priority is to advance along the reference path. The effect of this ratio is presented in section~\ref{sec: validation}.

Furthermore, the control velocity of the tool-tip (\ref{eq: LV_t}) could be represented with respect to any desired frame. Note that if the end-effector frame is selected, then the end-effector twist velocity $^e\mathbf{\underline{v}}_e$ is related to the linear velocity of the tool-tip $^e\mathbf{v}_t$ as
\begin{equation}	\label{eq: pf__v_t}
^e\mathbf{v}_t = \underset{\mathbf{L}_{pf} \in \mathbb{R}^{3 \times 6}}{\underbrace{ \left[ \mathbf{I}_{3\times3} ~~~ -[^e\mathbf{et}]_{\times} \right] }} \underset{^e\mathbf{\underline{v}}_e}{\underbrace{\left[ \begin{array}{c} ^e\mathbf{v}_e\\ ^e\mathbf{\omega}_e \end{array} \right]}}
\end{equation}
whereby $[^e\mathbf{et}]_{\times}$ is the skew-symmetric matrix associated to the vector $^e\mathbf{et}$, and $\mathbf{L}_{pf}$ is the interaction matrix related to the path-following task.

Finally, the control velocity for the path-following task is deduced as
\begin{equation}	\label{eq: taskPF}
^e\mathbf{\underline{v}}_e = \mathbf{L}_{pf}^\dagger ~^e\mathbf{v}_t \quad .
\end{equation}
%

\subsection{Bilateral Constrained Motion Controller}	\label{subSec: rcm}
As claimed above, the resection/ablation task is performed in a minimally invasive procedure. 
Therefore, the robot should perform the surgical task under the constraints of the incision point. This section begins with the description of RCM (bilateral constraints), while the following section describes the UCM (unilateral constraints). The RCM imposes that the center-line of tool body $\mathcal{S}_t$ should be coincident with the point $\mathcal{O}_{r}$. Simultaneously, the tool-tip must follow the desired path inside the incision orifice.

\begin{figure}[!h]
	\centering
	\includegraphics[width=0.65\columnwidth]{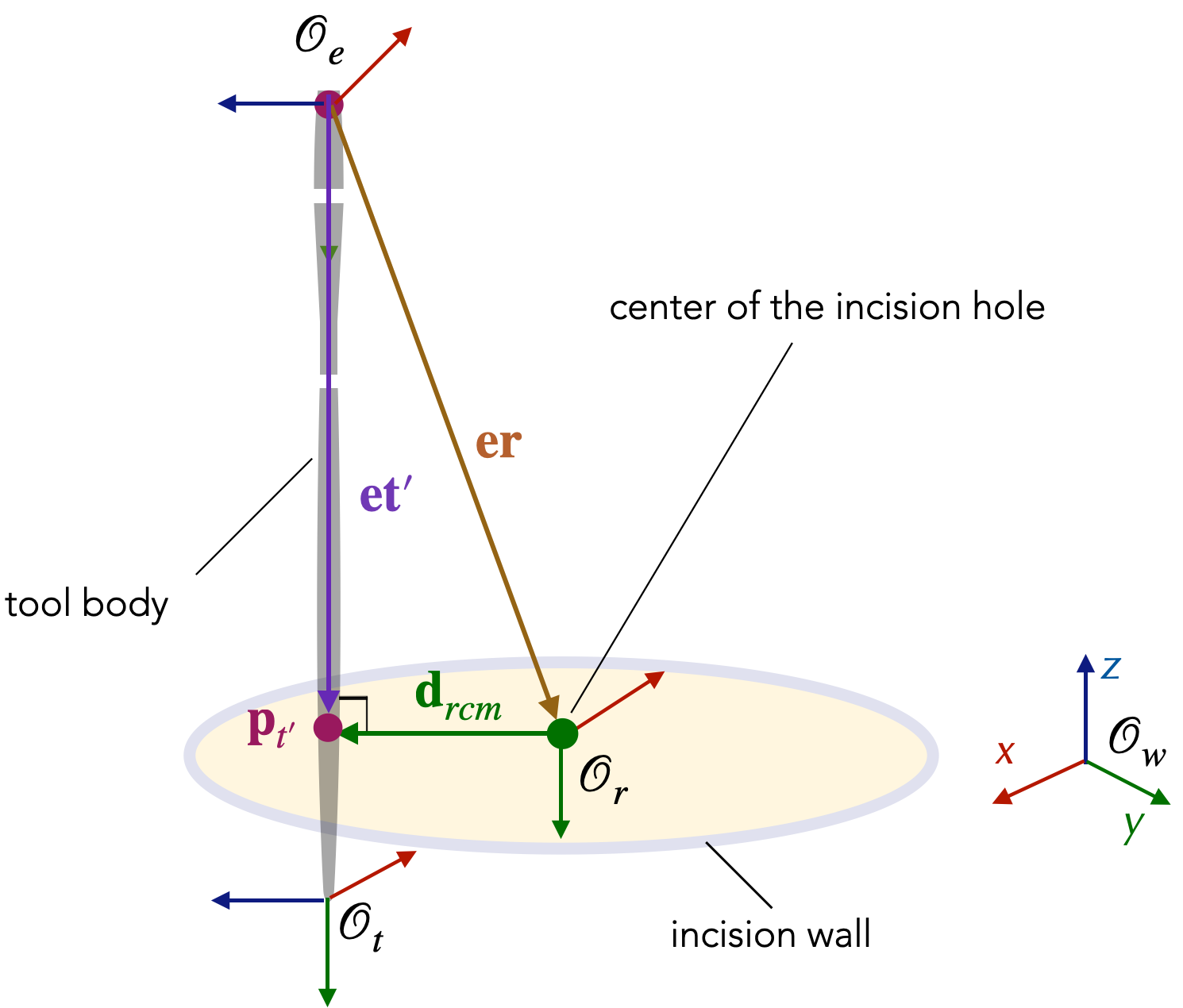}
	\caption{Geometric scheme of the bilateral linear error $\mathbf{d}_{rcm}$.}
	\label{fig: geomConcept_rcm}
\end{figure}
Fig.~\ref{fig: geomConcept_rcm} shows a straight tool which is located far from the center-point of incision orifice $\mathcal{O}_{r}$. 
The previous works~\cite{dahroug2017_vsPFRCM, dahroug2019_UCM} built the controller based on the angular error between the vectors $\mathbf{et}^\prime$ and $\mathbf{er}$ while the proposed controller in this section is based on the linear error $\mathbf{d}_{rcm}$. This new choice offers the controller to become independent of the tool shape. Let us imagine that the tool-tip position in Fig.~\ref{fig: geomConcept_rcm} is fixed in space, but its length can change. In the case of angular error, when the tool length increases, the error reduces its value. However, the linear error stays constant when the tool length changes. Therefore, the new choice grants better numerical~computing.

The error $\mathbf{d}_{rcm}$ is deduced by the orthogonal projection of the point $\mathcal{O}_{r}$ onto the tool body $\mathcal{S}_t$. The point $\mathbf{p}_{t^\prime}$ is resultant from the latter projection that is calculated as~follows
\begin{equation}	\label{eq: p_tPrime}
^e\mathbf{p}_{t^\prime} = ~^e\mathbf{u}_{et} ~^e\mathbf{u}_{et}^\top ~^e\mathbf{er}
\end{equation}
whereby $^e\mathbf{u}_{et}$ is the unit vector of $\mathbf{et}$ expressed in $\frame{e}$, and $^e\mathbf{er}$ represents the vector between both points $\mathcal{O}_{e}$ and $\mathcal{O}_{r}$ which is expressed in $\frame{e}$. 

In case the surgical tool is curved, the point $\mathbf{p}_{t^\prime}$ is determined by discretizing the tool body. Then the closest point onto the tool body is located. After that, the orthogonal projection is performed with respect to this point and the previous one on the tool center-line.
Thus, the error $\mathbf{d}_{rcm}$ is deduced~as
\begin{equation}	\label{eq: d_rcm}
\mathbf{d}_{rcm} = ~^e\mathcal{O}_{r} - ~^e\mathbf{p}_{t^\prime} 	\quad.
\end{equation}

The controller task is to find the spatial velocity of the robot end-effector $^e\mathbf{\underline{v}}_e$ for eliminating the rate-of-change of the bilateral linear error $\mathbf{d}_{rcm}$. Thereby, the time-derivative of the latter equation results in
\begin{equation}	\label{eq: dot__d_rcm}
\mathbf{\dot{d}}_{rcm} = ~^e\mathbf{v}_{r} - ~^e\mathbf{v}_{t^\prime} 
\end{equation}
where $^e\mathbf{v}_{t^\prime}$ is the linear velocity of the projected point $\mathbf{p}_{t^\prime}$ along the tool body, and $^e\mathbf{v}_{r}$ is the linear velocity of the trocar point described in $\frame{e}$. 
Indeed, the velocity of the projected point depends on the movement of the tool body with respect to the trocar point. Hence, this velocity is computed as~\cite{dahroug2019_UCM}
\begin{equation}		\label{eq: e_LV_t'}
^e\mathbf{v}_{t^\prime} = \frac{^e\mathbf{k}_t ~^e\mathbf{k}_{t}^{T}}{1 + \mathbf{d}_{rcm}^T (\mathbf{C}_t(s_t) \times ~^e\mathbf{k}_{t})} ~^e\mathbf{v}_{r}
\end{equation}
whereby $\mathbf{C}_t(s_t)$ is the tool curvature in the function of its arc length, and $^e\mathbf{k}_t$ is the instantaneous tangential unit-vector onto the tool curve/shape.

Since the calculation is done in the perspective of the end-effector frame $\frame{e}$, it implies that this frame is fixed, and the other ones are dynamic with respect to it. Consequently, the incision orifice virtually moves, and its linear velocity $^e\mathbf{v}_{r}$ is related to the spatial velocity of the robot end-effector thanks to the following formula 
\begin{equation}		\label{eq: e_LV_r}
^e\mathbf{v}_{r} = \underset{\mathbf{L}_{r} \in \mathbb{R}^{3 \times 6}}{\underbrace{\left[ \mathbf{I}_{3\times3} ~~~ -\left[^e\mathcal{O}_{r}\right]_\times \right]}} ~^e\mathbf{\underline{v}}_e 	\quad.
\end{equation}

By injecting the latter equation in (\ref{eq: e_LV_t'}) then the resultant in (\ref{eq: dot__d_rcm}), the time-derivative of the error $\mathbf{d}_{rcm}$ equals to
\begin{equation}	\label{eq: dot__d_rcm1}
\small
\mathbf{\dot{d}}_{rcm} = \underset{\mathbf{L}_{rcm} \in \mathbb{R}^{3 \times 6}}{\underbrace{
		\left[ \mathbf{I}_{3\time3} - \frac{^e\mathbf{k}_t ~^e\mathbf{k}_{t}^{T}}{1 + \mathbf{d}_{rcm}^T (\mathbf{C}_t(s_t) \times ~^e\mathbf{k}_{t})} \right]
		\left[ \mathbf{I}_{3\times3} ~~~ -\left[^e\mathcal{O}_{r}\right]_\times \right] }}
~^e\mathbf{\underline{v}}_e
\end{equation}
where $\mathbf{L}_{rcm}$ is the interaction matrix which relates between the end-effector velocity $^e\mathbf{\underline{v}}_e$ and the rate-of-change of the error $\mathbf{d}_{rcm}$.

Furthermore, a linearized proportional controller is applied to reduce the bilateral linear error in an exponential decay form. It defines the control velocity of the end-effector as
\begin{equation}	\label{eq: taskRCM}
^e\mathbf{\underline{v}}_e = - \lambda ~\mathbf{L}_{rcm}^{\dagger} ~\mathbf{d}_{rcm} .
\end{equation} 
whereby $\lambda$ is a positive gain which allows tuning the rate of exponential decay, and $\mathbf{L}_{rcm}^{\dagger}$ is the pseudo-inverse of the interaction matrix $\mathbf{L}_{rcm}$.

Finally, the RCM task can be combined as the highest priority with the path-following task as the secondary criteria. The hierarchical controller deduces the control velocity, by replacing the equations (\ref{eq: taskRCM}) and (\ref{eq: taskPF}) in equation (\ref{eq: th_generalForm2}), as
\begin{equation}	\label{eq: task_rcm_pf}
\begin{split}
^e\mathbf{\underline{v}}_e &= - \lambda \mathbf{L}_{rcm}^{\dagger}  \mathbf{d}_{rcm} + \mathbf{\tilde{L}}_{pf}^{\dagger} \left(  ~^e\mathbf{v}_t + \lambda \mathbf{L}_{pf} \mathbf{L}_{rcm}^{\dagger} \mathbf{d}_{rcm} \right), \\
&with \quad \mathbf{\tilde{L}}_{pf} = \mathbf{L}_{pf} \left( \mathbf{I} ~-~ \mathbf{L}_{rcm}^{\dagger} \mathbf{L}_{rcm} \right) \quad .
\end{split} 
\end{equation} 
In the opposite case, the hierarchical controller sets the path-following task (\ref{eq: taskPF}) as the highest priority while the RCM task (\ref{eq: taskRCM}) as the secondary one. The control velocity is deduced from equation (\ref{eq: th_generalForm2}) as
\begin{equation}	\label{eq: task_pf_rcm}
        ^e\mathbf{\underline{v}}_e = \mathbf{L}_{pf}^{\dagger} ~^e\mathbf{v}_t - \mathbf{\tilde{L}}_{rcm}^{\dagger} \left( \lambda \mathbf{d}_{rcm}  + \mathbf{L}_{rcm} \mathbf{L}_{pf}^{\dagger} ~^e\mathbf{v}_t \right), 
\end{equation} 
with
\begin{equation}	
    \mathbf{\tilde{L}}_{rcm} = \mathbf{L}_{rcm} \left( \mathbf{I} ~-~ \mathbf{L}_{pf}^{\dagger} \mathbf{L}_{pf} \right) \quad .
\end{equation} 
%

\subsection{Unilaterally Constrained Motion Controller}	\label{subSec: ucm}
This section continues with the design of the path-following controller under unilateral constraints. Notice that the UCM task assumes the incision orifice is larger than the tool diameter. Consequently, it imposes on the tool-tip to follow the incision/ablation path while the tool body is free to move within the incision orifice as long as it does not damage the orifice wall. Therefore, the formulation of the previous section needs to extend to satisfy the unilateral constraints.

\begin{figure}[!h]
	\centering
	\includegraphics[width=.7\columnwidth]{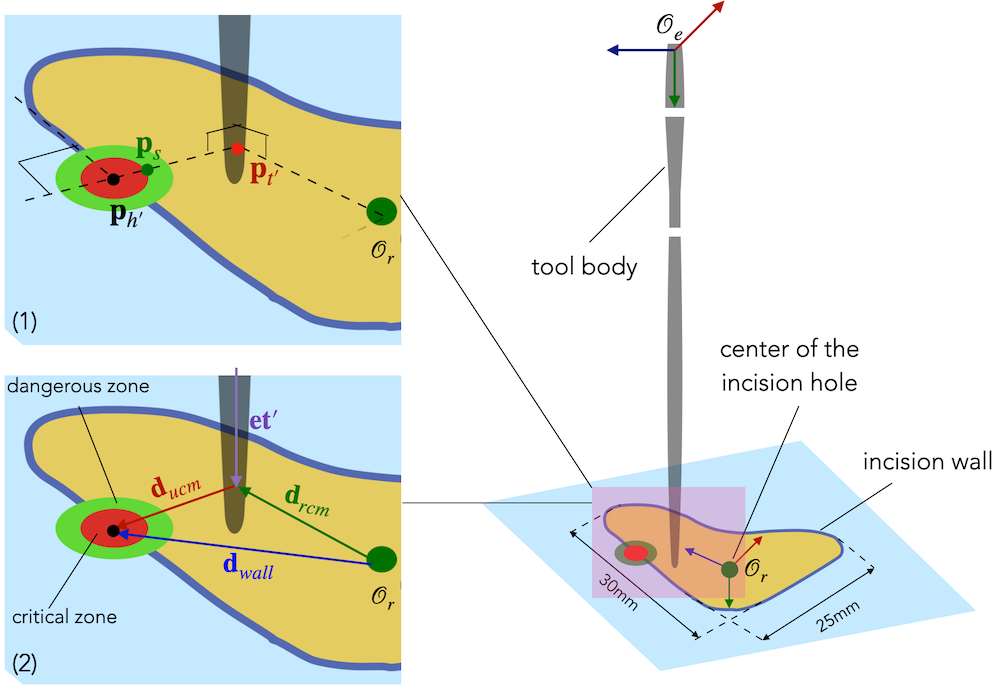}
	\caption{Geometric modelling of the unilateral linear error $\mathbf{d}_{ucm}$.}
	\label{fig: geomConcept_ucm}
\end{figure}
Fig.~\ref{fig: geomConcept_ucm}(left image 1) shows how the point $\mathbf{p}_{t^\prime}$ is orthogonally projected onto the orifice wall in order to determine the closest point $\mathbf{p}_{h^\prime}$ on the orifice wall $\mathcal{S}_{h}$. 
The distance between the latter two points forms the vector error $\mathbf{d}_{ucm}$ which can be defined as (left image 2 of Fig.~\ref{fig: geomConcept_ucm})
\begin{equation}	\label{eq: d_ucm}
\mathbf{d}_{ucm} = \underset{= \mathbf{d}_{rcm}}{\underbrace{^e\mathbf{t^\prime r}}}
- \underset{= \mathbf{d}_{wall}}{\underbrace{^e\mathbf{h^\prime r}}} .
\end{equation}

The question now is how to maintain the value of the error $\mathbf{d}_{ucm}$ greater or equal to~zero.
For security issues, three regions are defined around the projected point $\mathbf{p}_{h^\prime}$, as shown in the left image of Fig.~\ref{fig: geomConcept_ucm}: 
\begin{enumerate}
	\item \textit{critical zone} (dark red circle) which its border is defined by a minimal distance $d_{min}$;
	\item \textit{dangerous zone} (light green circle) which its border is defined by a maximal distance $d_{max}$; and 
	\item \textit{safe zone} which is the remain region outside the dangerous zone.
\end{enumerate}

When the Euclidean norm $\Vert \mathbf{d}_{ucm} \Vert$ is larger than the "dangerous" distance $d_{max}$, the tool can follow the reference path without any constraints since its location is in the safe zone. However, an admittance control is activated, which is composed of a virtual damper $\mu_{obs}$, when the tool body passes the dangerous zone border. Indeed, the admittance control imposes unilateral constraint towards the safe point $\mathbf{p}_{s}$ by generating a compensation velocity in the opposite direction to the orifice wall. 

By differentiating equation (\ref{eq: d_ucm}) with respect to time for deducing the velocity twist of the end-effector, it becomes equal to
\begin{equation}	\label{eq: dot__d_ucm}
\begin{split}
\mathbf{\dot{d}}_{ucm} &= \underset{\mathbf{\dot{d}}_{rcm}}{\underbrace{\left( ~^e\mathbf{v}_{r} - ~^e\mathbf{v}_{t^\prime} \right)}} - \underset{\mathbf{\dot{d}}_{wall}}{\underbrace{\left( ~^e\mathbf{v}_{r} - ~^e\mathbf{v}_{h^\prime} \right)}} \\
&= ~^e\mathbf{v}_{h^\prime} - ~^e\mathbf{v}_{t^\prime} \\
\end{split}
\end{equation}

The velocity of the projected point $\mathbf{p}_{h^\prime}$ is deduced in the same way as equation (\ref{eq: e_LV_t'}) 
\begin{equation}		\label{eq: e_LV_h'}
^e\mathbf{v}_{h^\prime} = \frac{^e\mathbf{k}_h ~^e\mathbf{k}_{h}^{T}}{1 + \mathbf{d}_{ucm}^T  \left( \mathbf{C}_h(s_h) \times ~^e\mathbf{k}_{h} \right) } ~^e\mathbf{v}_{t^\prime} \quad .
\end{equation}
where $\mathbf{C}_h(s_h)$ is the orifice curvature in function of its arc length, and $^e\mathbf{k}_h$ is the instantaneous tangential unit-vector onto the orifice curve.
In another perspective, the latter equation describes how the projection of the point $\mathbf{p}_{t^\prime}$ onto the geometric curve of the orifice wall $\mathcal{S}_h$ evolves with time. 

The velocity $^e\mathbf{v}_{t^\prime}$ is deduced by combining equations (\ref{eq: e_LV_t'}) and (\ref{eq: e_LV_r}) 
\begin{equation}		\label{eq: e_LV_tPrime2}
^e\mathbf{v}_{t^\prime} = \underset{\mathbf{L}_{v_{t^\prime}} \in \mathbb{R}^{3 \times 6}}{\underbrace{
		\frac{^e\mathbf{k}_t ~^e\mathbf{k}_{t}^{T}}{1 + \mathbf{d}_{rcm}^T \left( \mathbf{C}_t(s_t) \times ~^e\mathbf{k}_{t} \right) } 
		\left[ \mathbf{I}_{3\times3} \quad -\left[^e\mathcal{O}_{r}\right]_\times \right] }} 
~^e\mathbf{\underline{v}}_e     \quad .
\end{equation}

Replacing equations (\ref{eq: e_LV_h'}) and (\ref{eq: e_LV_tPrime2}) in (\ref{eq: dot__d_ucm}) yields
\begin{equation}	\label{eq: dot__d_ucm2}
\mathbf{\dot{d}}_{ucm} = \underset{\mathbf{L}_{ucm} \in \mathbb{R}^{3 \times 6}}{\underbrace{ 
		\left( \frac{^e\mathbf{k}_h ~^e\mathbf{k}_{h}^{T}}{1 + \mathbf{d}_{ucm}^T \left( \mathbf{C}_h(s_h) \times ~^e\mathbf{k}_{h} \right)} -  \mathbf{I}_{3\times3} \right)  \mathbf{L}_{v_{t^\prime}} }}
~^e\mathbf{\underline{v}}_e     
\end{equation}
whereas $\mathbf{L}_{ucm}$ is the interaction matrix that relates the twist end-effector with the rate of change of the error $\mathbf{d}_{ucm}$.

Thereby, the control velocity of the UCM task is defined as
\begin{equation}	\label{eq: taskUCM}
^e\mathbf{\underline{v}}_e = - \mu_{obs} \lambda \mathbf{L}_{ucm}^\dagger \mathbf{d}_{ucm}  \quad .
\end{equation}

The damping coefficient $\mu_{obs}$ changes following a sigmoid function that depends on the vector $\mathbf{d}_{ucm}$. It means that the gain $\mu_{obs}$ reaches its minimal value when $\mathbf{d}_{ucm}$ is higher than the safe distance $d_{max}$, where the tool location in the dangerous zone. However, $\mu_{obs}$ gradually increases until it reaches its maximal value when $\mathbf{d}_{ucm}$ is smaller than the critical distance $d_{min}$, where the tool location in the critical zone. This behaviour is modeled as 
\begin{equation} 	\label{eq: mu_obs} 
\mu_{obs} = \dfrac{\sigma_{max}}{1 + e^{\Big( \sigma_{step} \big( \Vert \mathbf{d}_{ucm} \Vert - \sigma_{min} \big) \Big)}}   
\end{equation}
where $\sigma_{max}$, $\sigma_{min}$ and $\sigma_{step}$ are tunable parameters for modifying the sigmoid form.

Finally, the path-following task can be combined as the highest priority with the UCM task as the secondary criteria. The hierarchical controller deduces the control velocity, by replacing the equations~(\ref{eq: taskUCM}) and~(\ref{eq: taskPF}) in equation~(\ref{eq: th_generalForm2}), as
\begin{equation}	\label{eq: task_ucm_pf}
    \begin{split}
        ^e\mathbf{\underline{v}}_e &= \mathbf{L}_{pf}^{\dagger}  ~^e\mathbf{v}_{t} - \mathbf{\tilde{L}}_{ucm}^{\dagger} \left( \mu_{obs} \lambda \mathbf{d}_{ucm} + \mathbf{L}_{ucm} \mathbf{L}_{pf}^{\dagger} ~^e\mathbf{v}_{t} \right), \\
        &with \quad \mathbf{\tilde{L}}_{ucm} = \mathbf{L}_{ucm} \left( \mathbf{I} ~-~ \mathbf{L}_{pf}^{\dagger} \mathbf{L}_{pf} \right) \quad .
    \end{split} 
\end{equation} 
%

\section{VALIDATION} \label{sec: validation}
This section discusses several scenarios to evaluate qualitatively and quantitatively the proposed methods and materials.
The developed controllers were first tested using our simulator framework and then in an experimental set-up that takes up the various components of the simulator.

\begin{figure}[!h]
	\centering
	\includegraphics[width=\columnwidth]{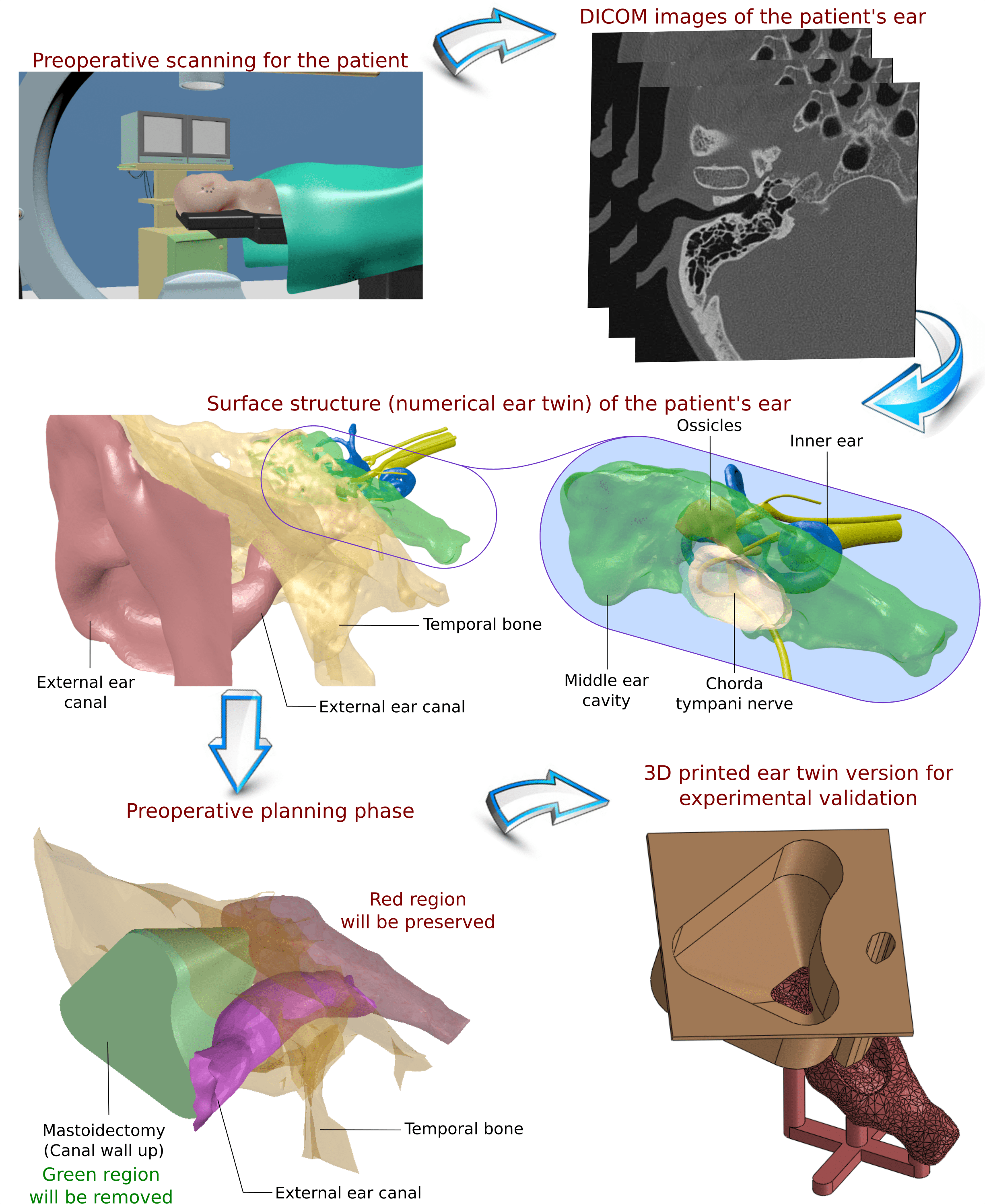}
	\caption{The steps done to achieve a numerical and physical model of the middle ear~cavity.}
	\label{fig: stepsEarModel}
\end{figure}
%
\subsection{Implementation Issues}\label{subSec: implementation}
This part begins by converting the patient's ear to its numerical-twin and then its 3D printed-twin. The first step to accomplish this job is the scan of the patient's ear during the preoperative phase for getting DICOM (Digital Imaging and Communications in Medicine) images, as depicted in Fig.~\ref{fig: stepsEarModel}. The DICOM images are handled by the software \textit{3D Slicer} which converts these images to a 3D surface model after a segmentation process. Prior works were done in relation to this subject for achieving an automated segmentation process (e.g., \cite{noble2009automaticSeg, fauser2019}). However, the segmentation process that we have done manually is not automated since this is not the focus of this article. In the future, we believe that our segmentation process needs to be done again in an automated manner for efficiency. 

The 3D Slicer software exports the segmentation results as \textit{STL} files for each anatomical structure. Afterward, the software \textit{MeshLab} treats the STL files for smoothing the surface and reducing the number of vertices and faces to cut down the final STL file size. This step produces the numerical-twin of the patient's ear.

The next step creates the 3D printed-twin for conducting the experimental validation. Indeed, a simplified version of the numerical-twin is imported in \textit{Solidworks} for i) adding some thickness to the middle ear cavity and ii) creating the incision orifice through the mastoid. 

After that, the planning stage of the desired path within the middle ear cavity begins. The path planning step can be optimized (e.g., \cite{kazemi2010path, fauser2018planning}). 
However, this step was done manually on Solidworks to generate text files that contain the geometry of the reference path and the orifice wall as a sequence of 3D points. These files are inputs for the controller. 
This step should be investigated in the future and add to the adequate functions in the simulator.

\begin{figure}[!btp]
	\centering
	\includegraphics[width=\columnwidth]{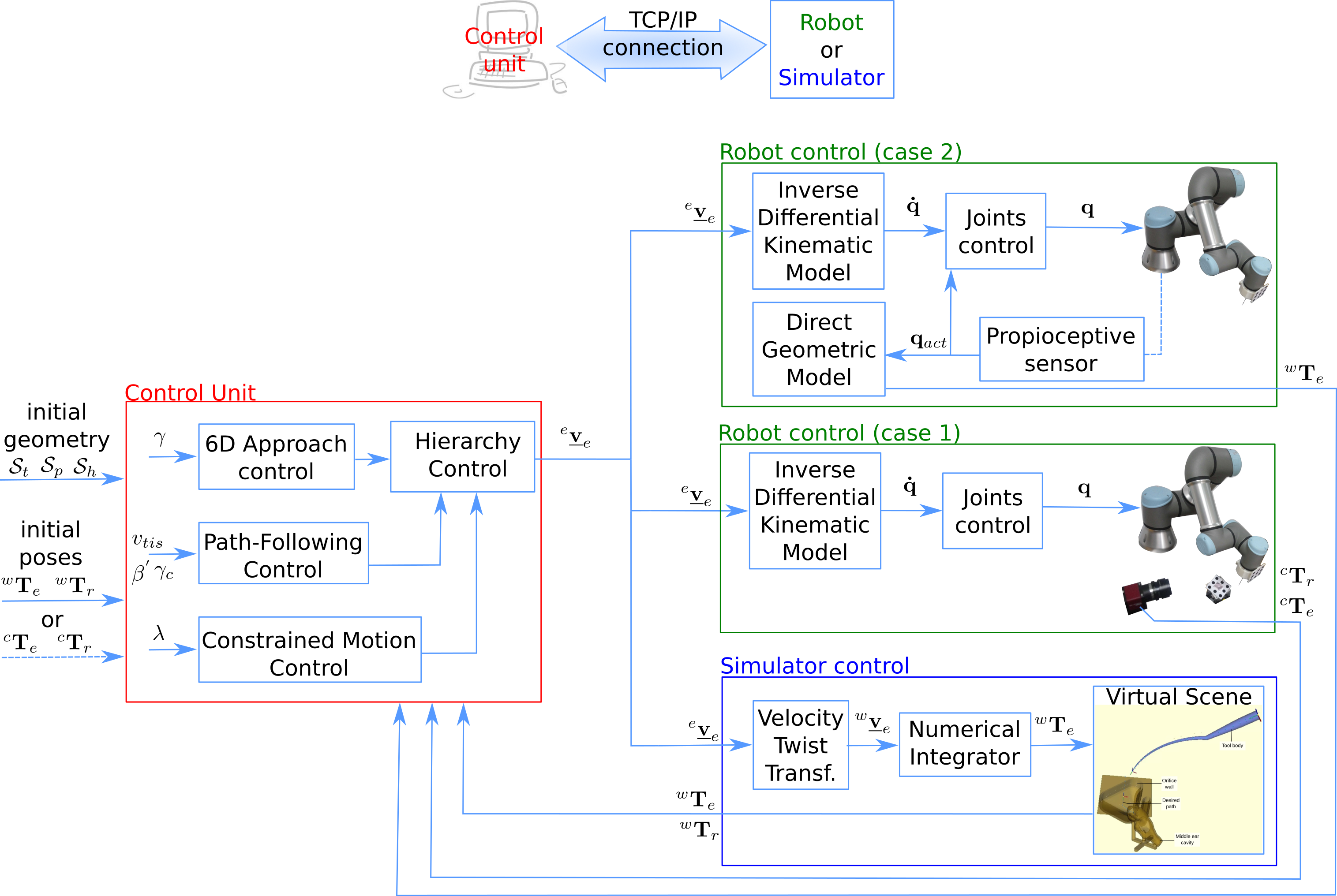}
	\caption{Block diagram of the TCP/IP communication between the client (proposed controller) and the server (simulator or robot) or vice-versa.}
	\label{fig: blockDiagramControl}
\end{figure}
Fig.~\ref{fig: blockDiagramControl} presents the proposed control architect with the TCP/IP communication. This architect allows easy interchangeability between the real-system (robot) and its numerical-twin (simulator). The latter figure (the red block at the left-hand side) also shows that the implemented controller is firstly initialized with the end-effector and the incision orifice poses, $^\star\mathbf{T}_e$ and $^\star\mathbf{T}_r$ respectively. These poses must be described in the same frame (e.g., the world frame $\frame{w}$ or the camera $\frame{c}$). Indeed, the tool geometry $\mathcal{S}_t$ is defined with respect to the end-effector frame $\frame{e}$ while the reference path $\mathcal{S}_p$ and the orifice wall $\mathcal{S}_h$ are described in the incision orifice frame $\frame{r}$. 
Furthermore, the controllers should be initialized by the different gain coefficients before the control-loop starts. 

The hierarchy controller arranges throughout the control-loop the priority between the different tasks (i.e., the approach task, the path-following task, and the RCM/UCM constraints). 
Indeed, the control-loop is mainly divided into three phases: 
\begin{enumerate}
	\item the \textit{outside phase}: the tool corrects its initial pose with respect to the incision orifice. This stage applies the approach task for regulating: i) the tool-tip position to the point located before the orifice center point, and ii) the tool-tip rotation as the rotation of the orifice reference frame. This manoeuvrer is performed to ensure some security for the next phase;
	\item the \textit{transition phase}: the tool-tip passes the center point of the incision orifice. The RCM controller could oscillate when the trocar point is close to the tool-tip. These oscillations are generated because the controller computes large rotation displacement, due to the lever phenomena, for compensating the rotation error. Thus, the trocar point is virtually moved to the first point on the reference path. Consequently, the tool body can rotate about this new point. This virtual trocar point moves towards the orifice frame while the tool-tip advances along the reference path;
	\item the \textit{inside phase}: the tool-tip follows the desired path while the tool body is constrained by the orifice wall or the orifice center point.
\end{enumerate}
Therefore, the output of this block is the spatial velocity of the end-effector expressed in its frame ($^e\mathbf{\underline{v}}_e$) while its inputs are the instantaneous poses of the end-effector and the incision orifice ($^\star\mathbf{T}_e$ and $^\star\mathbf{T}_r$). 
The question now is: what is the observation frame? 

In the simulator case (the blue block at the right-hand side of Fig.~\ref{fig: blockDiagramControl}), it is straightforward since the user initializes the poses with respect to the world frame $\frame{w}$ of the virtual scene. Thus, the spatial velocity $^e\mathbf{\underline{v}}_e$ is transformed to $^w\mathbf{\underline{v}}_e$ then it is integrated over the sample time $T_{e}$ to deduce the new pose of the end-effector. Consequently, the tool pose is updated in the virtual scene, and this new pose is sent back to the control unit block for computing a new iteration. 

There are two options for designing the control architect in the experimental case. 
The first one consists of using an exteroceptive sensor (e.g., camera) for estimating the required poses. This option is depicted in the green block of Fig.~\ref{fig: blockDiagramControl} named \textit{Robot control (case 1)}. The input of this block is the spatial velocity $^e\mathbf{\underline{v}}_e$ that is transformed to deduce the angular velocity of each joint $\mathbf{\dot{q}}$ with the help of the inverse differential kinematic model to move mechanical structure of the robot. This motion is observed from the camera frame $\frame{c}$ in order to estimate the new pose of the end-effector and that of the orifice. These poses are the output of this block which are sent back to the control unit block for calculating a new iteration. However, this option is uneasy for implementation since it needs a particular setup to accurately track both the end-effector and the orifice~\cite{gerber2014}. 

The second option is more fundamental than the first one. It is also presented in the green block of Fig.~\ref{fig: blockDiagramControl} named \textit{Robot control (case 2)}. It uses the proprioceptive sensors of the robot and its forward geometric model to estimate the end-effector pose. Despite that, this option requires performing a registration process~\cite{cleary2010review, gerber2014} between the robot and the orifice before the control-loop. After that, the robot works blindly, and the user assumes that the orifice does not move during the control-loop. 

The simulator is implemented in C++. It uses \textit{Eigen} library for linear algebra (e.g., vectors, matrices, numerical solvers) and \textit{PCL} (Point Cloud Library) for visualizing the STL parts and converting them to point clouds. This conversion is done to initialize the collision detection that is accomplished by \textit{VCollide} library. Finally, \textit{ViSP} library is used for manipulating the camera images throughout the experimental work.

\subsection{Numerical Validation}
A numerical simulator was developed, as the first step, to validate the functioning of the diverse methods before physical implementation. It simulates the geometric motion of the surgical tool through the incision orifice and the middle ear cavity. 
The software interchangeability of the simulator and the physical set-up allowed us also to tune the controller parameters before the experimental validation. 
Therefore, this part presents three scenarios for the demonstration:
\begin{itemize}
	\item \textit{scenario 1} performs the path-following task without any constraint applied on the tool motion. It demonstrates the effect of the gain coefficients $v_{tis}$ and $\beta$ in equations~(\ref{eq: choice__LV_t}) and (\ref{eq: choice__beta}), respectively, on the performance of the path-following controller;
	\item \textit{scenario 2} performs the path-following task with RCM constraints. It simulates the drilling of a minimal invasive tunnel (i.e., conical tunnel) through the mastoid portion to reach the middle ear cavity;
	\item \textit{scenario 3} assumes the surgeon performed a standard mastoidectomy. It simulates an inspection/resection task performed under the UCM constraints.
\end{itemize}
%

\subsubsection{Simulation of the path-following task without constraints}
Throughout this first trial, the value of $v_{tis}~=~4mm/second$ in equation~(\ref{eq: choice__LV_t}) remains constant during all tests. 
Besides that, the same reference path is tested during this trial, and it is defined as a spiral curve. 

\begin{figure}[!h]
	\centering
	\includegraphics[width=.7\columnwidth]{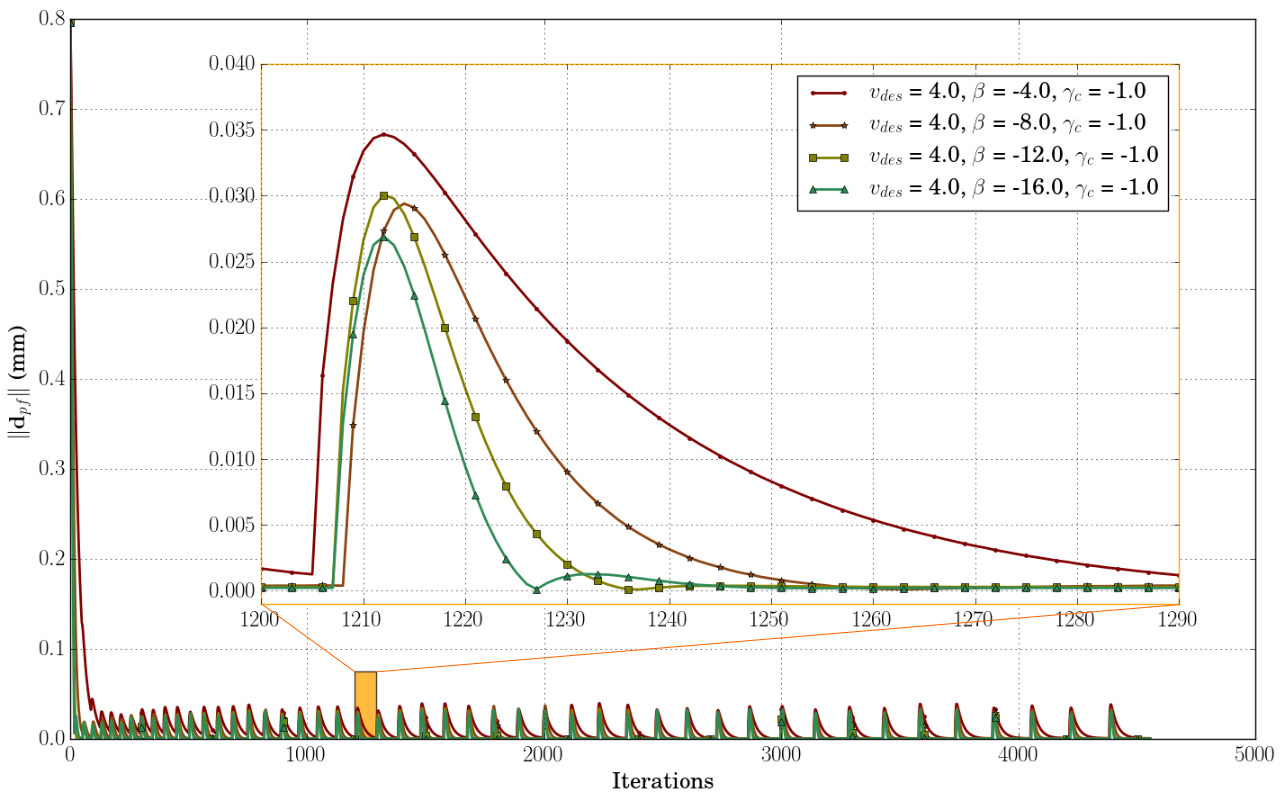}
	\caption{The effect of the ratio between $v_{des}$ and $\beta^{\prime}$ on the path-following error $\mathbf{d}_{pf}$ with a zoom and magnification on the orange region.}
	\label{fig: error_d_pf__changing_beta}
\end{figure}

The first group of tests keeps the value of $\gamma_c$ in equation~(\ref{eq: choice__beta}) constant while decreasing the value of $\beta^{\prime}$ which its value varies from $-4$ to $-16$.
Fig.~\ref{fig: error_d_pf__changing_beta} shows the influence of the gain coefficient $\beta^{\prime}$ on the path-following error $\mathbf{d}_{pf}$. Indeed, this error computed as in equation~(\ref{eq: d_pf}). 
The ripples appearing in this figure represent the linear error between the projected point $\mathbf{p}_{t^\prime}$ and the closest point on the reference path $\mathbf{p}_{p^\prime}$. An orange rectangle appeared in this figure for zooming on one of these ripples. One can observe that the error reduced as designed exponentially.

The latter figure also demonstrates that the best ratio between $\beta^{\prime}$ and $v_{tis}$ should be greater than $-2$ (the saddle-brown line with star markers), and less than or equal $-3$ (the olive line with square markers). If the ratio is less than or equal to $-1$, the controller response is relatively slow, and there is a steady-state error (the maroon line with round markers in Fig.~\ref{fig: error_d_pf__changing_beta}). On the opposite, if the ratio is higher than or equal to $-4$, the system begins to oscillate (having over-shoots). However, the controller reduces the error faster than the previous cases (the sea-green line with triangular markers in Fig.~\ref{fig: error_d_pf__changing_beta}).

The second group of tests chose a constant ratio $-2$ while decreasing the value of $\gamma_{c}$ from $-2$ to $-16$. This group shows that the best value of $\gamma_c$ is to be near from $\beta^{\prime}$. If $\gamma_c$ is higher than $\beta^{\prime}$, the system begins to have over-shoots, but it reduces faster the path-following error.

\subsubsection{Simulation of a robotic drilling task under RCM constraint}
The surgeon perforates manually until now the mastoid portion in the temporal bone for reaching the middle ear cavity. The resultant mastoidectomy orifice is invasive. Thereby, a less invasive tunnel is proposed in this trial. Besides that, the drilling procedure becomes automated so that the surgeon can concentrate on other essential tasks. Indeed, this drilling procedure is achieved by merging the approach task, the 3D path-following task, and the RCM task.

\begin{figure}[!h]
	\centering
	\subfloat[]{
		\includegraphics[width=0.45\columnwidth]{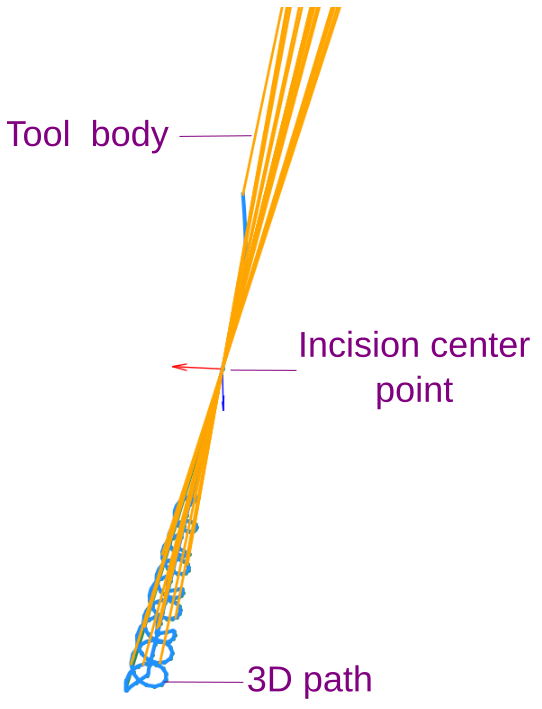}
		\label{fig: sim_pf_rcm_st_dp_motion}
	}
	\hfil
	\subfloat[]{
		\includegraphics[width=0.45\columnwidth]{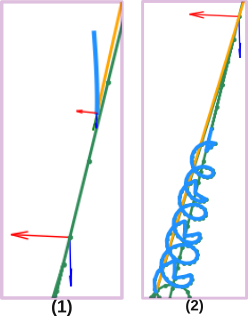}
		\label{fig: sim_pf_rcm_st_dp_zoomMotion}
	}
	\caption{Numerical validation of the 3D path-following under a RCM constraint (see Extension 2). (a) The tool pose with respect to the desired path. (b) Sequence of zoom images during the tool motion.}
	\label{fig: sim_pf_rcm_straightTool_drillPath}
\end{figure}
Fig.~\ref{fig: sim_pf_rcm_straightTool_drillPath} depicts the tool motion throughout the drilling procedure. 
The subplot (a) draws the tool geometry and its poses at different instances (orange straight-lines). It also shows the drilling path defined as a combination of spiral and linear portions (sea-green dotted-line). One can view that the tool body is always coincident with the orifice center point.
The subplot (b1) shows the path done by the tool-tip (dodger-blue line) to accomplish the outside phase by i) approaching towards the point located before the orifice center point, and ii) regulating the rotation of the tool-tip frame to be as that of the orifice reference frame.
The subplot (b2) depicts an instantaneous zoom on the tool pose during the inside phase to visualize the RCM~effect.

%
\begin{figure}[!h]
	\centering
	\includegraphics[width=.7\columnwidth]{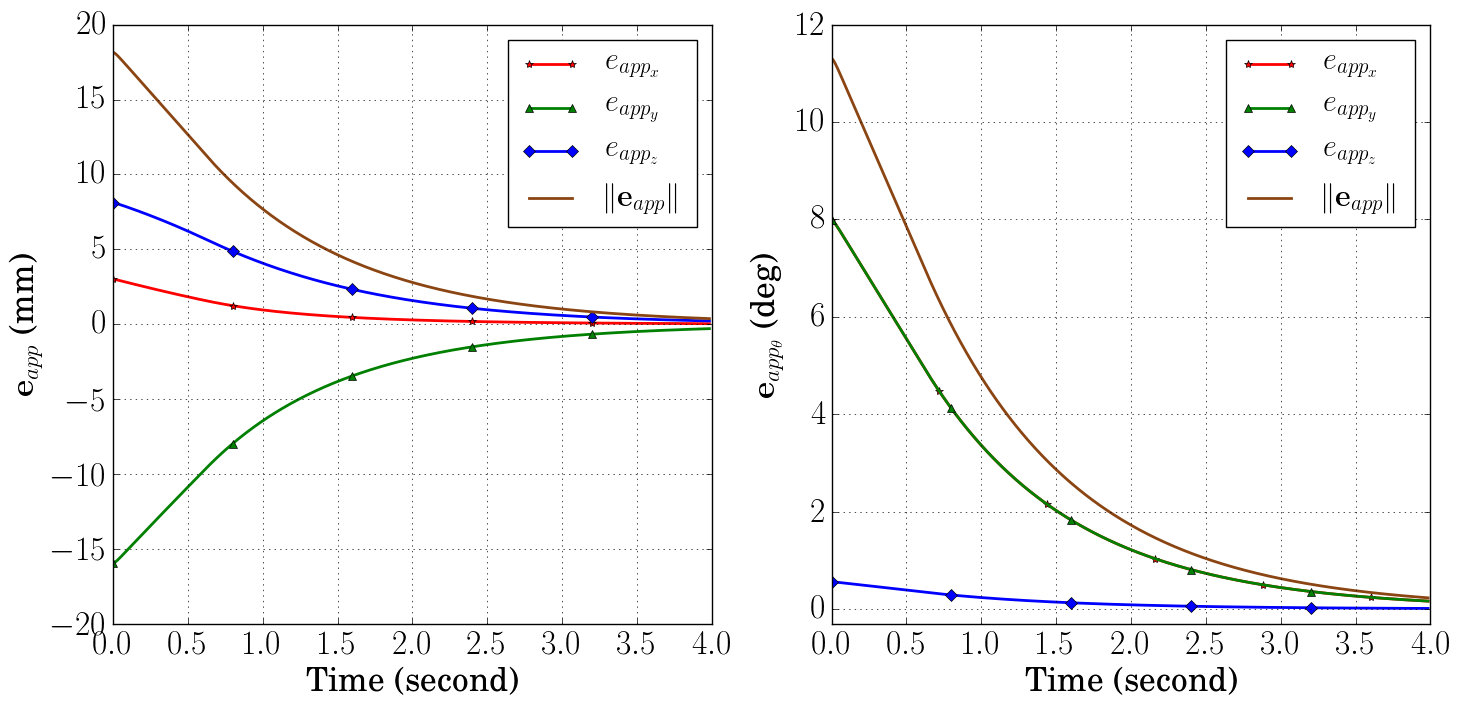}
	\caption{The approach task error $\mathbf{e}_{app}$, where the left column is the linear error and the right column represents the angular error.}
	\label{fig: sim_rcm_st_dp_error__e_app}
\end{figure}
\begin{figure}[!h]
	\centering
	\includegraphics[width=\columnwidth]{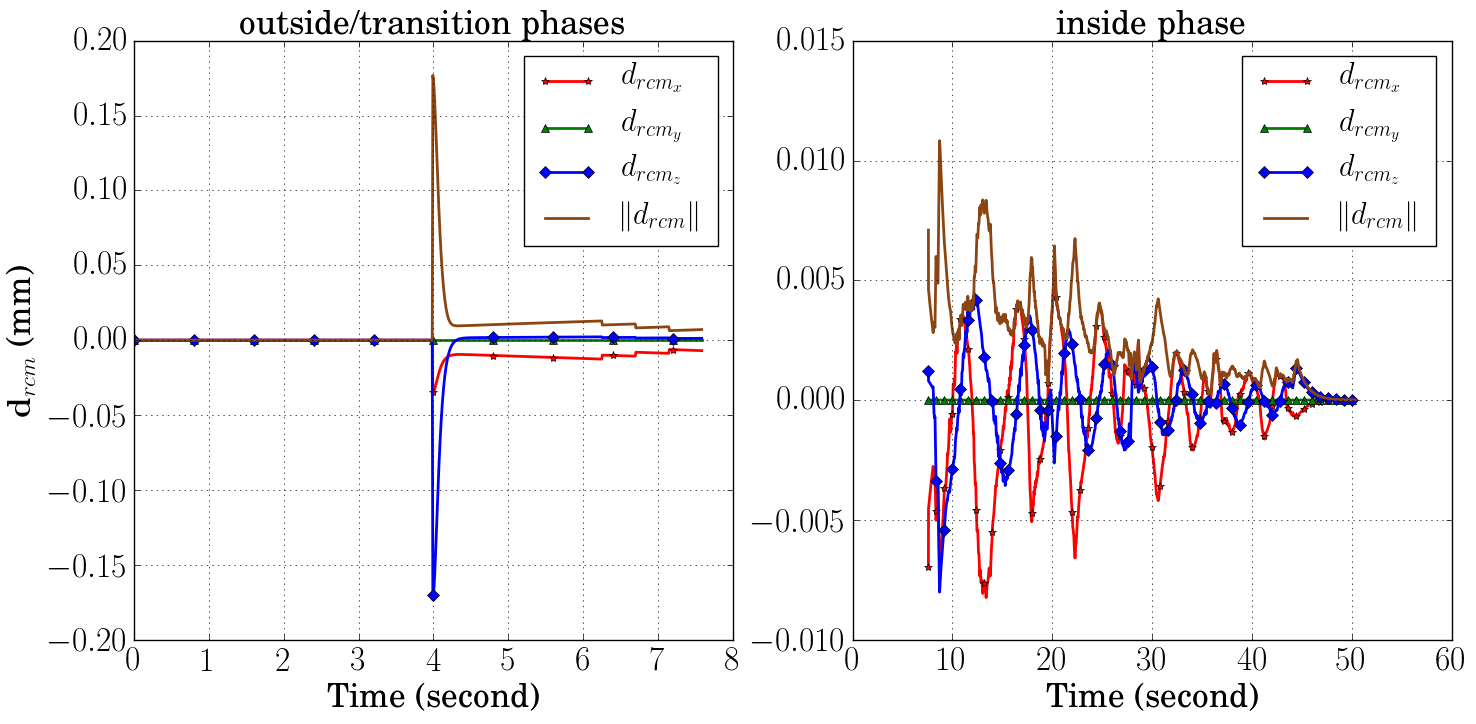}
	\caption{The RCM task error $\mathbf{d}_{rcm}$, where the left column shows the error evolution during the transition phase while the right column presents the error during the inside~phase.}
	\label{fig: sim_rcm_st_dp_error__d_rcm}
\end{figure}
\begin{figure}[!h]
	\centering
	\includegraphics[width=\columnwidth]{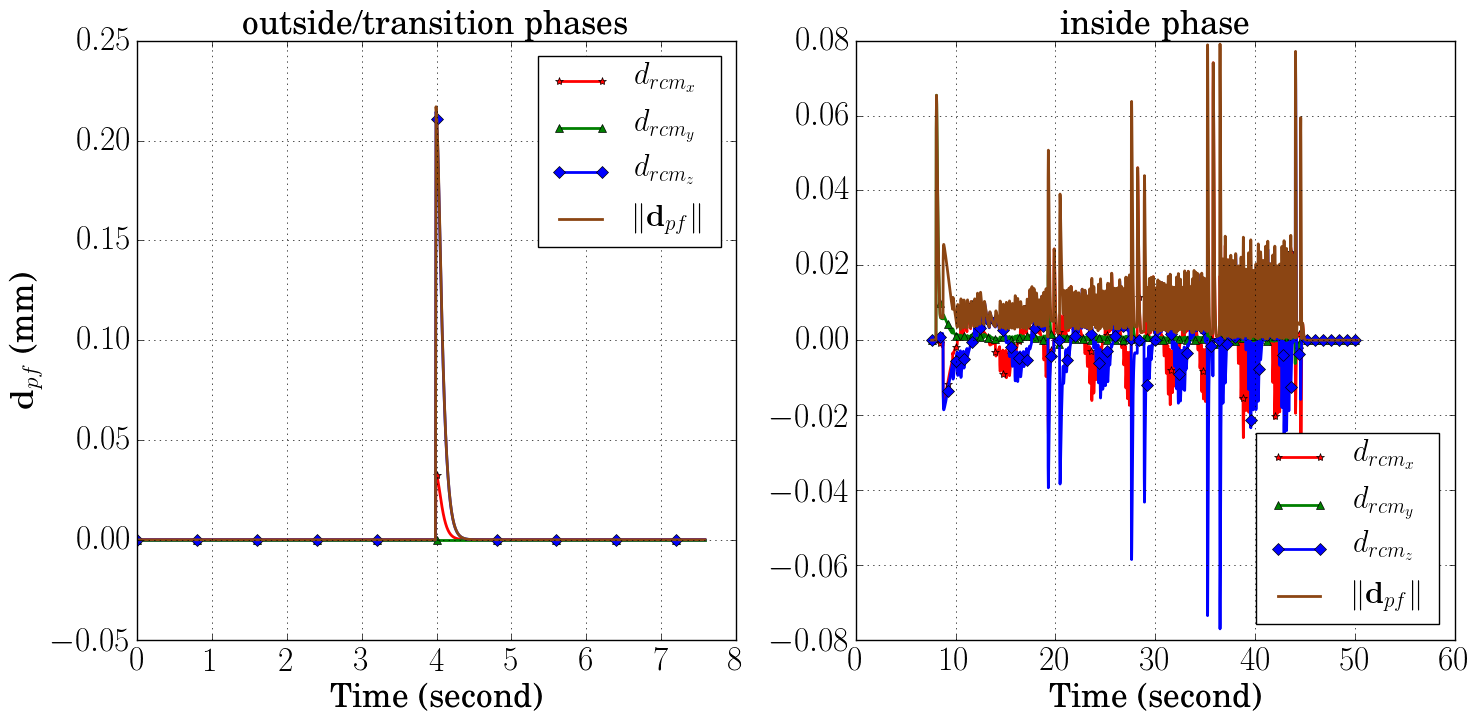}
	\caption{The path-following task error $\mathbf{d}_{pf}$, where the left column shows the error evolution during the transition phase while the right column presents the error during the inside phase.}
	\label{fig: sim_rcm_st_dp_error__d_pf}
\end{figure}
%

The approach task error $\mathbf{e}_{app}$ computed in equation~(\ref{eq: controlApp}) is visualized in Fig.~\ref{fig: sim_rcm_st_dp_error__e_app} which depicts the linear errors in the column and the angular errors in the right one. Over this period, the error is reduced in an exponential form as planned.

At the end of the latter period, the transition phase starts.
The task-hierarchical controller becomes active, and it arranges the path-following task as the highest priority while the RCM task is the second one. The errors of these tasks presented in the left columns of Fig.~\ref{fig: sim_rcm_st_dp_error__d_pf} and \ref{fig: sim_rcm_st_dp_error__d_rcm} which are obtained from equations (\ref{eq: d_pf}) and (\ref{eq: d_rcm}) for the path-following and RCM errors, respectively. One can observe a peak appeared around $4~seconds$ in the path-following figure due to the initial error when the controller becomes activated. Then, it attenuates the error until it attains stability. 
Furthermore, one can visualize in the RCM figure that three peaks appeared at the end of this phase. This behaviour happened due to the movement of the virtual trocar point.

After the previous period, the inside phase starts where the hierarchical controller modifies the priority by setting the RCM task as the highest one while the path-following is the secondary one.
The RCM task error $\mathbf{d}_{rcm}$ was computed as $0.002\pm0.002~mm$ (mean error $\pm$ STD (STandard Deviation) error), as shown in the right column of Fig.~\ref{fig: sim_rcm_st_dp_error__d_rcm}, while the path-following error $\mathbf{d}_{pf}$ was $0.008\pm0.009~mm$, as shown in the right column of Fig.~\ref{fig: sim_rcm_st_dp_error__d_pf}.
The gain values used for this trail were equal to $\lambda = 1$, $\gamma = 1$, $v_{tis} = 4~mm/second$, $\beta^\prime = -10$, $\gamma_{c} = -10$ and $T_e = 0.008~second$.

\subsubsection{Simulation of an ablation/excision surgical task under UCM constraint}
\begin{figure}[!h]
	\centering
	\subfloat[]{
		\includegraphics[width=0.37\columnwidth]{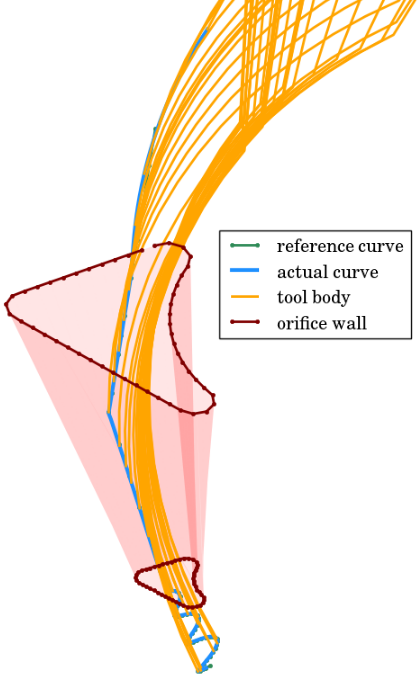}
		\label{fig: sim_pf_ucm_ct_mp_motion}
	}
	\hfil
	\subfloat[]{
		\includegraphics[width=0.4\columnwidth]{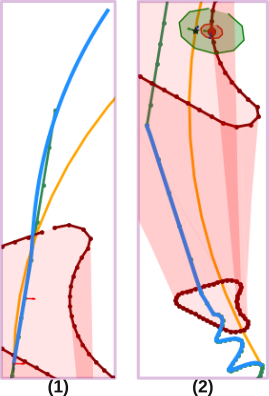}
		\label{fig: sim_pf_ucm_ct_mp_zoomMotion}
	}
	\caption{Numerical validation of the 3D path-following under a UCM constraint (see Extension 3). (a) The tool pose with respect to the desired path. (b) Sequence of zoom images during the tool motion.}
	\label{fig: sim_pf_ucm_curvedTool_mastoidPath}
\end{figure}

In this trial, the incision orifice size is larger than the instrument diameter. The tool is consequently subject to the UCM for providing more freedom to the tool movements inside the incision orifice. 
This behaviour is shown in Fig.~\ref{fig: sim_pf_ucm_ct_mp_motion} where the orifice wall is represented by the red surface. 
The latter figure also presents the curved tool employed during this trial which performs an ablation or scanning process.
The desired 3D path is thus composed of a linear portion to reach the middle ear cavity and a spiral curve to simulate the required surgical task.
This selected path can reach some regions where a straight tool cannot attain (see Extension 4 to visualize the collision of the latter one with the orifice~wall).

\begin{figure}[!h]
	\centering
	\includegraphics[width=.7\columnwidth]{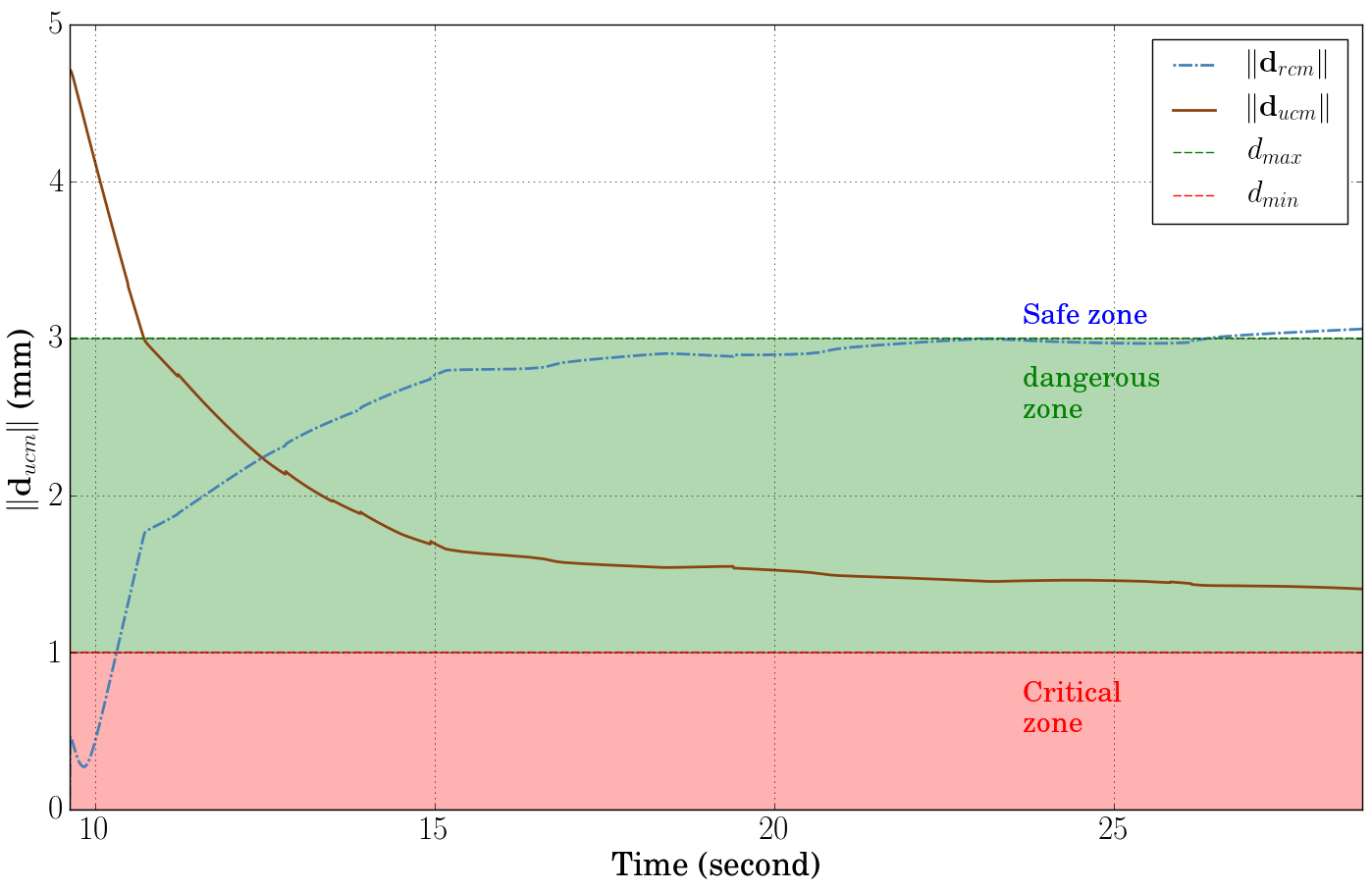}
	\caption{The UCM task error $\mathbf{d}_{ucm}$ during the inside phase along side the error $\mathbf{d}_{rcm}$.}
	\label{fig: sim_ucm_ct_mp_error__d_ucm}
\end{figure}
\begin{figure}[!h]
	\centering
	\includegraphics[width=\columnwidth]{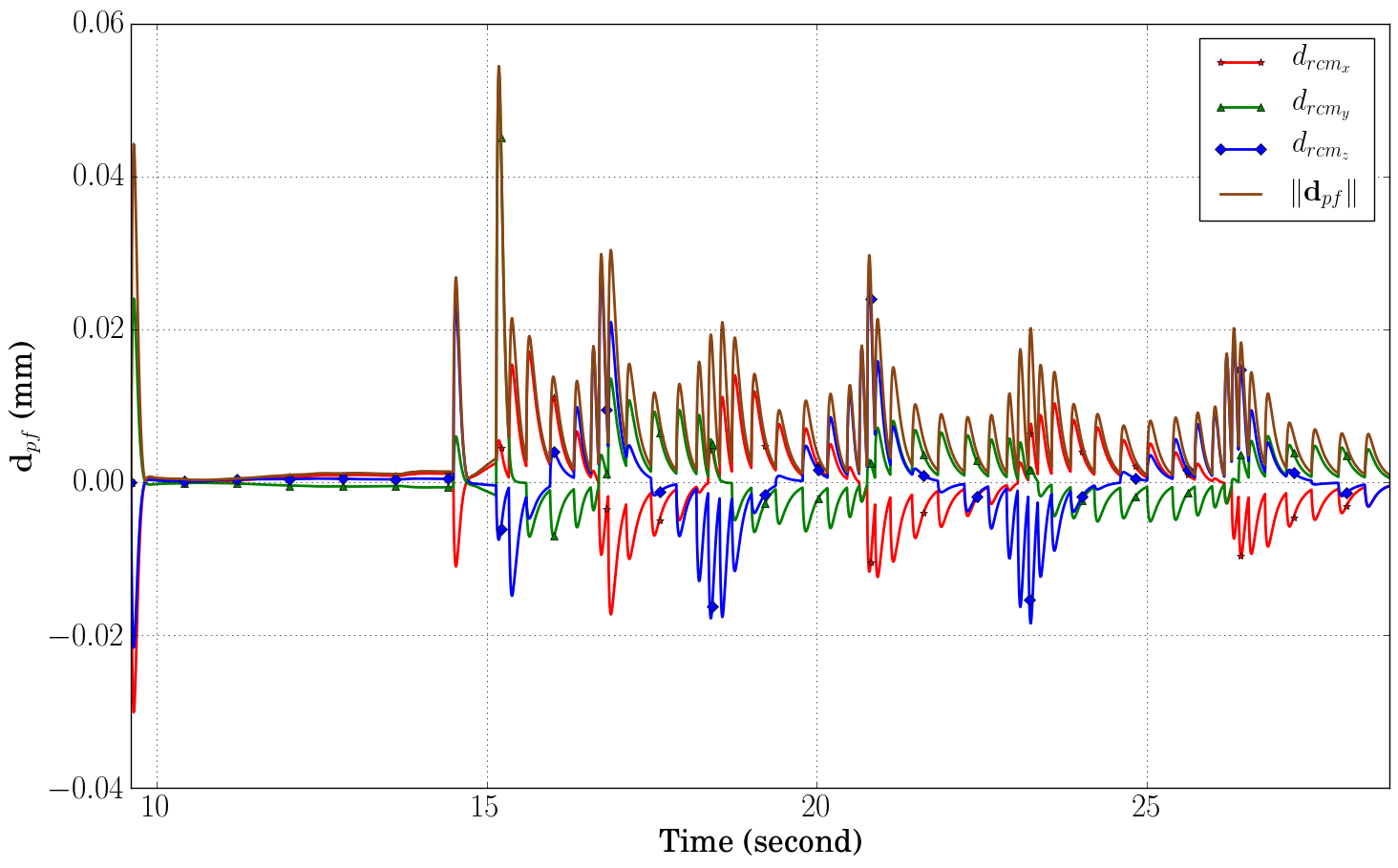}
	\caption{The path-following task error $\mathbf{d}_{pf}$ during the inside phase.}
	\label{fig: sim_ucm_ct_mp_error__d_pf}
\end{figure}

The subplot (b1) of Fig.~\ref{fig: sim_pf_ucm_ct_mp_zoomMotion} indicates the path done by the tool during the outside phase. It also presents an instantaneous pose of the tool body throughout the transition phase. As explained in the previous trial, the proposed controller executes the same tasks over these two phases.
Subplot (b2) presents the tool motion during the inside phase, where the dangerous and critical zones are represented by the green and red circles, respectively. The center point of these circles corresponds to the point $\mathbf{p}_{h^\prime}$ obtained by projecting $\mathbf{p}_{t^\prime}$ onto the orifice wall~$\mathcal{S}_{h}$.

Throughout the inside phase, the hierarchical controller combines the UCM task with the path-following task as described in~(\ref{eq: task_ucm_pf}). Fig.~\ref{fig: sim_ucm_ct_mp_error__d_ucm} shows the UCM task error  $\mathbf{d}_{ucm}$ which is deduced as in equation~(\ref{eq: d_ucm}). It also presents the boundaries of the critical and dangerous zones. One can observe that the error $\mathbf{d}_{ucm}$ begins with a considerable value, compared to the error $\mathbf{d}_{rcm}$, since the previous phase delivers the tool to the center point of the incision orifice. Then, the error $\mathbf{d}_{ucm}$ reduced, while the error $\mathbf{d}_{rcm}$ increased because the tool approached the incision wall to follow the reference path. However, the error $\mathbf{d}_{ucm}$ did not exceed the $d_{min}$, which implies the tool body did not enter the critical zone.

Fig.~\ref{fig: sim_ucm_ct_mp_error__d_pf} presents the path-following error $\mathbf{d}_{pf}$ during the inside phase. It was measured was $0.005\pm0.006~mm$. The gain values used for this trail were equal to $\lambda~=~0.8$, $\gamma~=~0.8$, $v_{tis}~=~4~mm/second$, $\beta^\prime~=~-10$, $\gamma_{c}~=~-10$ and $T_e~=~0.008~second$.

\subsection{Experimental Validation}
\begin{figure}[!h]
	\centering
	\includegraphics[width=.9\columnwidth]{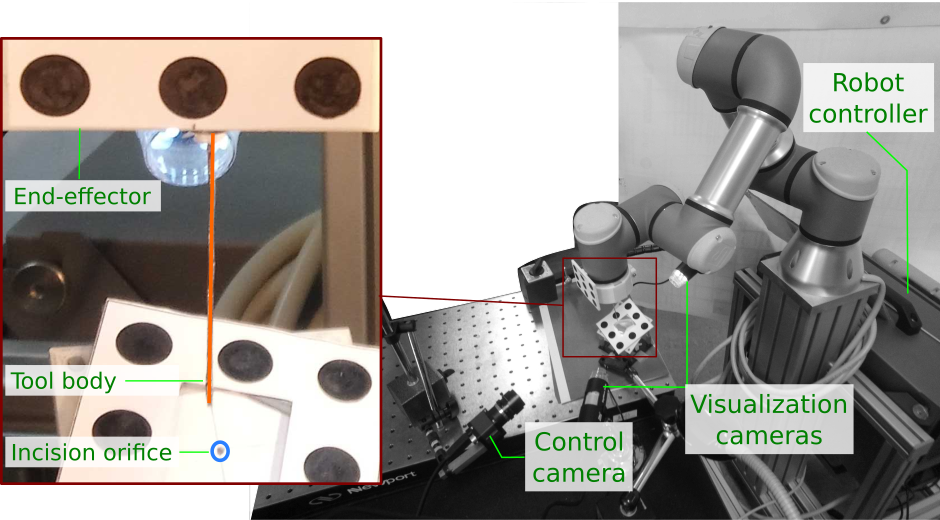} 
	\caption{Configuration of the experimental setup.}
	\label{fig: manipNotation}
\end{figure}

This part is devoted to the physical implementation of the blocks \textit{Robot control} that is shown in Fig.~\ref{fig: blockDiagramControl}. Its physical correspondence is presented in Fig.~\ref{fig: manipNotation}.
The robotic work-cell in the latter figure consists of:
\begin{itemize}
	\item a serial robot from \textit{Universal Robot} (UR3) with $\pm 0.03~mm$ pose repeatability. It communicates with the proposed controller via TCP/IP for receiving the command velocity of the end-effector. It also sends the end-effector pose to the controller if the block \textit{Robot control (case 1)} is required to be executed;
	\item a monocular camera from \textit{Guppy} (with image size $640\times420~pixels$) and an optical objective lens from \textit{Computar} with distortion (model MLM3X-MP) are used for the control purpose. This optical system tracks and estimates the poses of the end-effector and the incision orifice. It then sends these poses to the proposed controller if the block \textit{Robot control (case 2)} is needed to be executed;
	\item two visualization cameras provide other views for recording the multimedia~videos.
\end{itemize}
%
%

%
%
%
\begin{figure}[!btp]
	\centering
	\subfloat[]{
		\includegraphics[width=0.6\columnwidth]{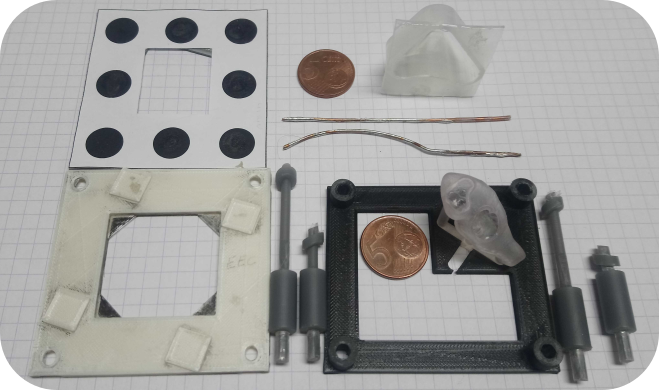}
		\label{fig: exp_earModelParts}
	}
	\subfloat[]{
		\includegraphics[width=0.4\columnwidth]{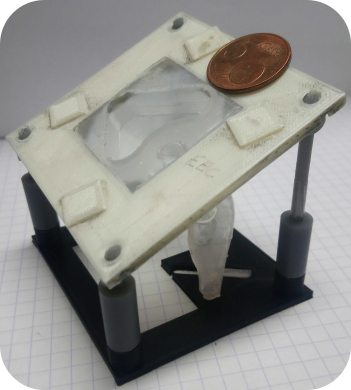}
		\label{fig: exp_earModelAssem}
	}
	\caption{The printed ear model used during the different tests. (a) The different parts of the ear model and the rigid tools. (b) After assembling the different parts.}
	\label{fig: exp_earModel}
\end{figure}

The numerical twin of the ear model shown previously in Fig.~\ref{fig: stepsEarModel} is modified for implementing its 3D printed twin. This modification holds up the mastoidectomy orifice with the middle ear cavity and a planar grid/marker. Fig.~\ref{fig: exp_earModel} presented the fabricated parts before and after the assembly, alongside the rigid tools used during the validation tests.

The trials of this part have the objective to evaluate the performance of the path-following controller under constraints. Therefore, a curved tool follows the same planned path, one time under the RCM constraint and the second time under the UCM constraint.

\subsubsection{Path-Following under RCM Constraint}	
\begin{figure}[!h]
	\centering
	\subfloat[]{
		\includegraphics[width=0.35\columnwidth]{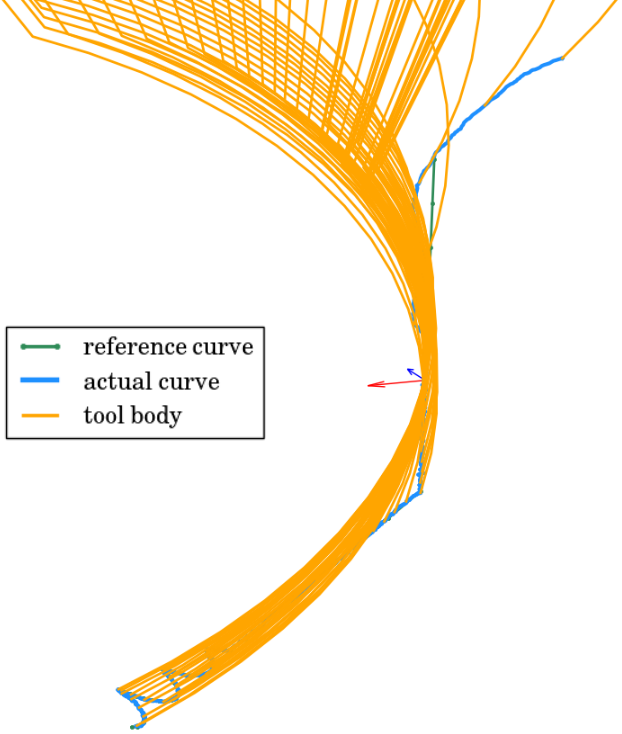}
		\label{fig: exp_pf_rcm_ct_mp_motion}
	}
	\hfil
	\subfloat[]{
		\includegraphics[width=0.58\columnwidth]{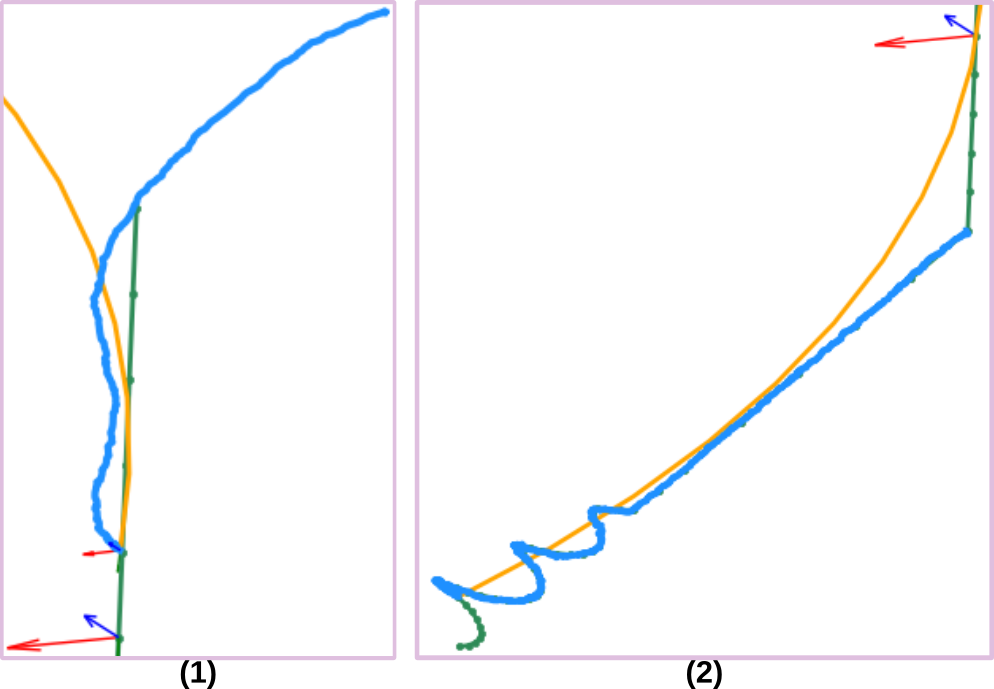}
		\label{fig: exp_pf_rcm_ct_mp_zoomMotion}
	}
	\caption{Experimental validation of the 3D path-following under a RCM constraint (see Extension 5). (a) The tool pose with respect to the desired path. (b) Sequence of zoom images during the tool motion.}
	\label{fig: exp_pf_rcm_curvedTool_mastoidPath}
\end{figure}

\begin{figure}[!h]
	\centering
	\includegraphics[width=0.9\columnwidth]{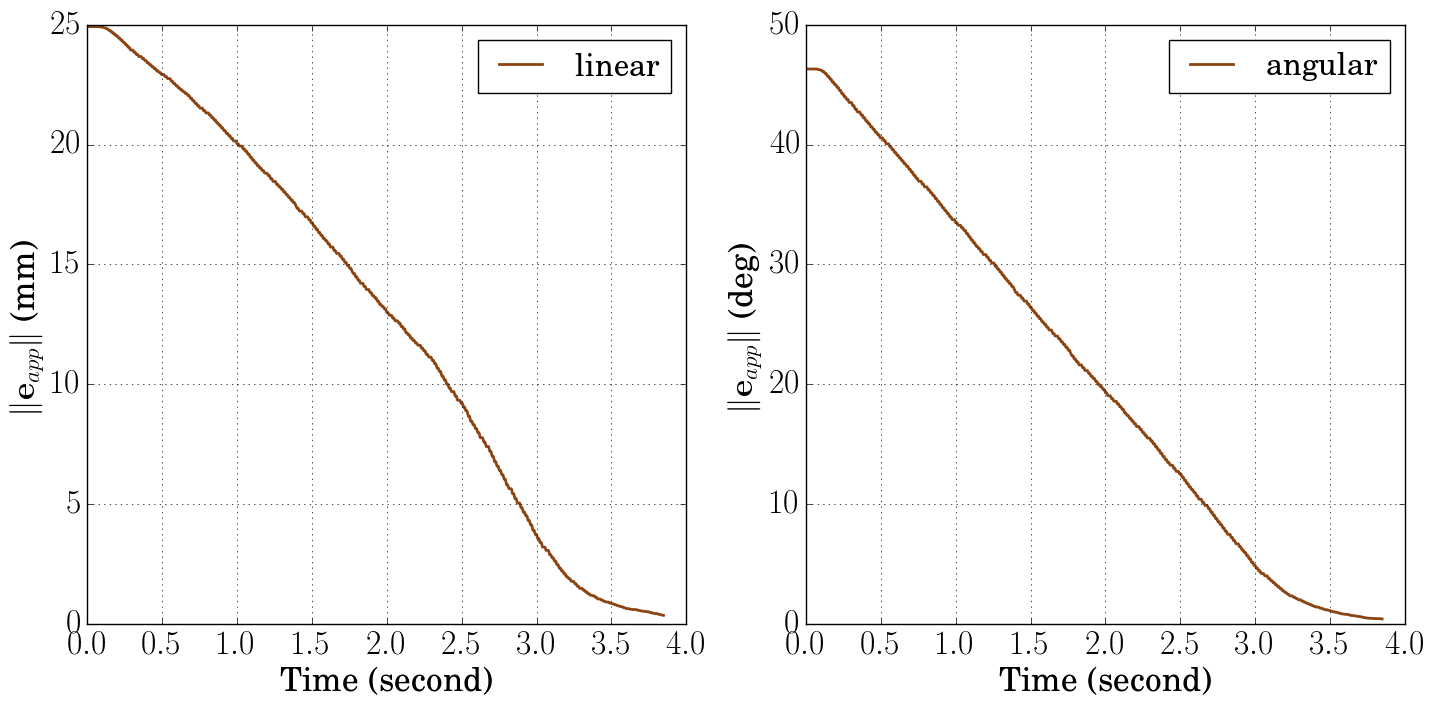}
	\caption{The approach task error $\mathbf{e}_{app}$, where the left column is the linear error and the right column represents the angular error.}
	\label{fig: exp_rcm_ct_mp_error__e_app}
\end{figure}

Fig.~\ref{fig: exp_pf_rcm_curvedTool_mastoidPath} presents the desired path (sea-green dotted line), the resultant motion of the curved tool (orange line), and the path done by tool-tip (dodger-blue line). 
One can observe in Fig.~\ref{fig: exp_pf_rcm_ct_mp_zoomMotion}(1) that the tool approaches to the incision orifice by executing the controller given in equation~(\ref{eq: controlApp}). The approach task error $\mathbf{e}_{app}$ computed from equation~(\ref{eq: approachError}). Fig.~\ref{fig: exp_rcm_ct_mp_error__e_app} presents the latter error and it converges toward zero by the end of this phase.

\begin{figure}[!h]
	\centering
	\includegraphics[width=\columnwidth]{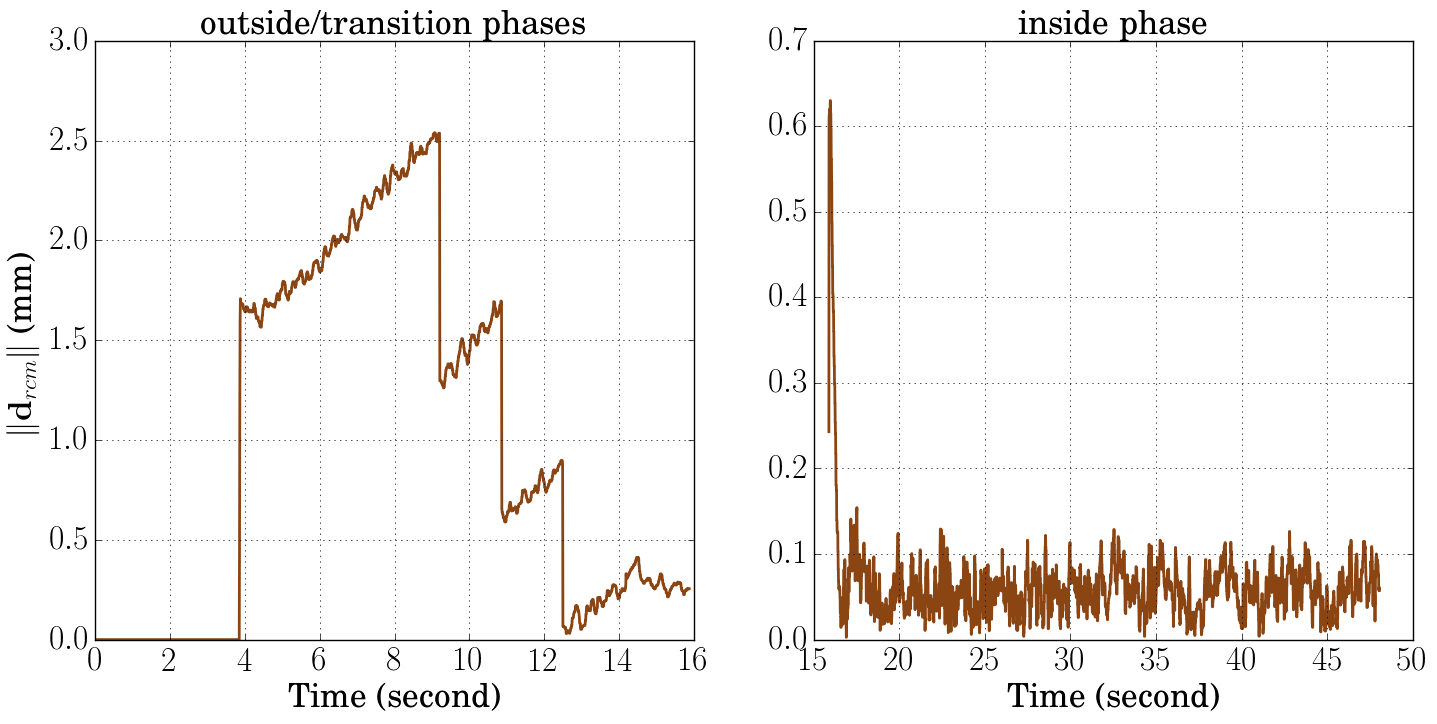}
	\caption{The RCM task error $\mathbf{d}_{rcm}$, where the left column shows the error evolution during the outside/transition phases while the right column presents the error during the inside phase.}
	\label{fig: exp_rcm_ct_mp_error__d_rcm}
\end{figure}

\begin{figure}[!h]
	\centering
	\includegraphics[width=\columnwidth]{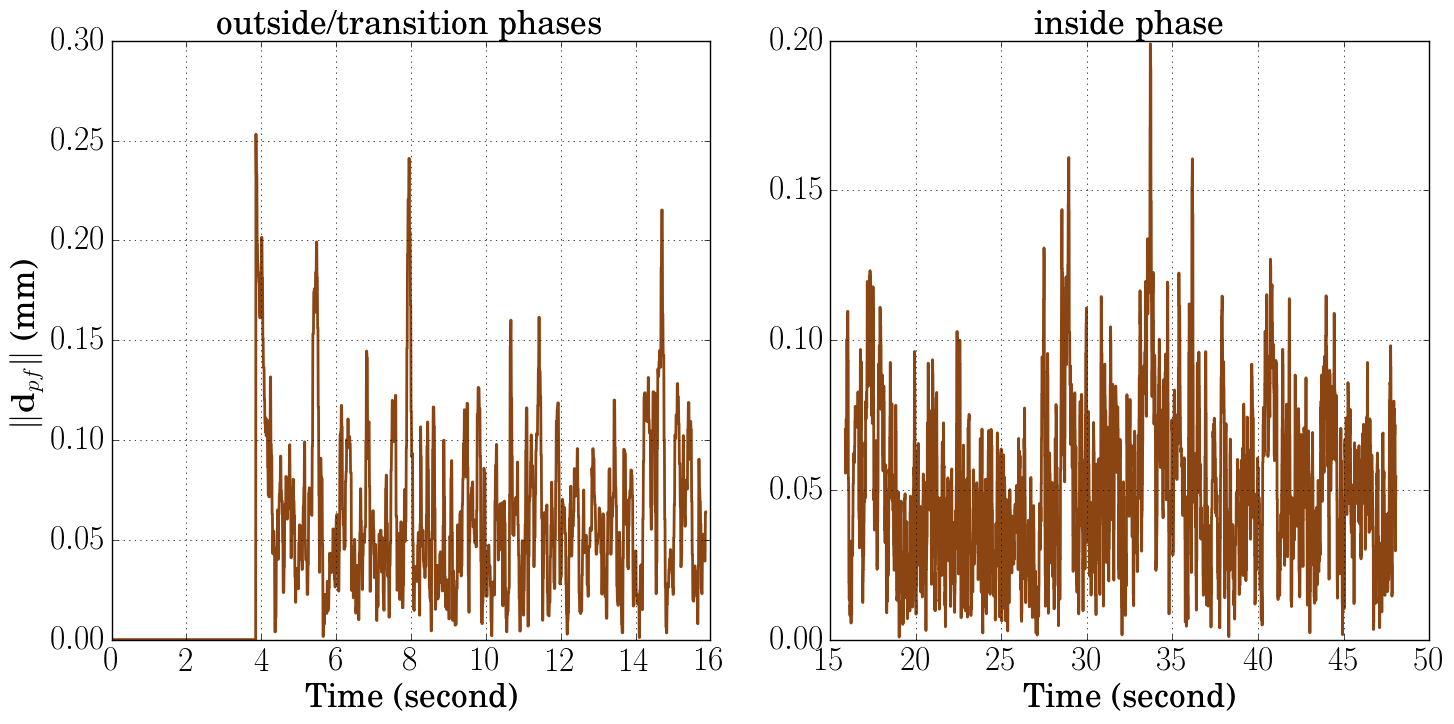}
	\caption{The path-following task error $\mathbf{d}_{pf}$, where the left column shows the error evolution during the outside/transition phases while the right column presents the error during the inside phase.}
	\label{fig: exp_rcm_ct_mp_error__d_pf}
\end{figure}

Afterward, the transition phase starts so that the tool passes the center point of the incision orifice, as explained previously. 
The hierarchical controller (equation~\ref{eq: task_pf_rcm}) arranges the path-following task as the highest priority while the RCM task is the second one. 
This behaviour is demonstrated in the left column of Fig.~\ref{fig: exp_rcm_ct_mp_error__d_rcm}-\ref{fig: exp_rcm_ct_mp_error__d_pf}, where the hierarchical controller has been activated around $4~second$. One can visualize that the RCM task error $\mathbf{d}_{rcm}$ has some steps due to the movements of the virtual trocar point while the path-following error $\mathbf{d}_{pf}$ maintained its value around zero.


When the tool passes the center point of the incision orifice, the inside phase begins.
The hierarchical controller (equation~\ref{eq: task_rcm_pf}) modifies its priorities by setting the RCM task as the highest one and the path-following as the second one. The system performances during the inside phase are shown in the right columns of Fig.~\ref{fig: exp_rcm_ct_mp_error__d_rcm}-\ref{fig: exp_rcm_ct_mp_error__d_pf}. 
During this phase, the RCM task error $\mathbf{d}_{rcm}$ measured as $0.06 \pm 0.05 mm$ (mean error $\pm$ standard deviation (STD) error) while the path-following error $\mathbf{d}_{pf}$ was $0.05 \pm 0.03 mm$. 

A exteroceptive sensor used to close the control loop, as presented in Fig.~\ref{fig: blockDiagramControl} by the block \textit{Robot control (case 2)}. Besides that, the gain values used in this experiment were equal to  $\lambda = 1$, $\gamma = 1$, $v_{tis} = 0.5~mm/second$, $\beta^\prime = -1.25$, $\gamma_{c} = -10$ and $T_e = 0.008~second$.

Another trial was conducted for testing the block \textit{Robot control (case 1)} by using the proprioceptive sensor in the control loop. The system performances are better than the exteroceptive test (see test 2 in Table~\ref{table: expTests}). The errors $\mathbf{d}_{rcm}$ and  $\mathbf{d}_{pf}$ are reduced to almost half. It implies that our vision system needed amelioration in terms of accuracy.

From the surgeon's perspective, it is required to target the residual cells of cholesteatoma. It implies that the robot should detect/remove a human cell whose size is around $0.1mm$. The proposed controller reached the requirements since the error $\mathbf{d}_{pf}$ is smaller than the human cell size. Besides that, the surgical tool does not damage the entry orifice (patient's~head).

By increasing the tool velocity $v_{tis} = 2~mm/second$ and maintain the same ratio $\beta^\prime/v_{tis} = -2$, the system performances deteriorated as expected. The errors $\mathbf{d}_{rcm}$ and $\mathbf{d}_{pf}$ are almost increase by half (see tests 2 and 4 in Table~\ref{table: expTests}). Therefore, the choice of the gain coefficients effects the system~performances.

\subsubsection{Path Following under UCM Constraint}	
\begin{figure}[!h]
	\centering
	\subfloat[]{
		\includegraphics[width=0.52\columnwidth]{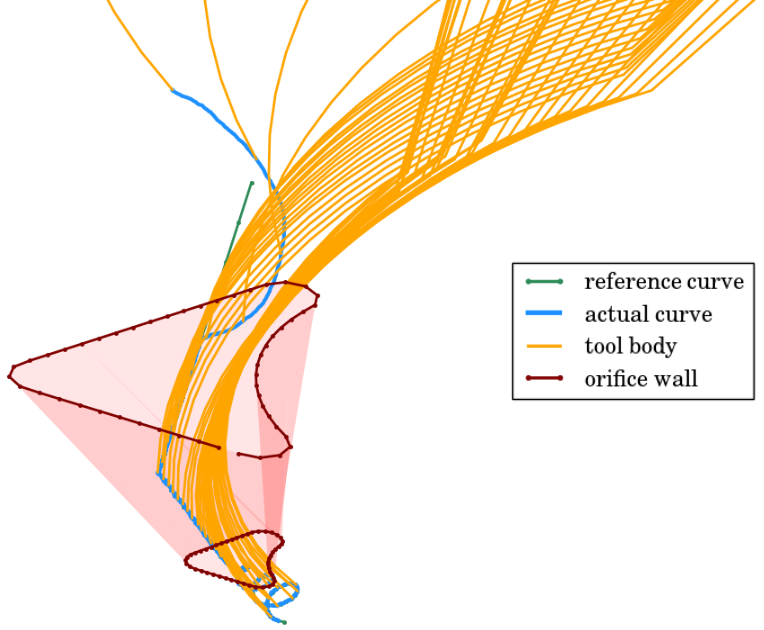}
		\label{fig: exp_pf_ucm_ct_mp_motion}
	}
	\hfil
	\subfloat[]{
		\includegraphics[width=0.41\columnwidth]{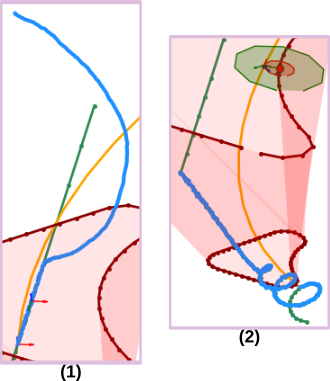}
		\label{fig: exp_pf_ucm_ct_mp_zoomMotion}
	}
	\caption{Experimental validation of the 3D path-following under a UCM constraint (see Extension 6). (a) The tool motion during the different phases. (b) Sequence of zoom images during the tool motion. }
	\label{fig: exp_pf_ucm_curvedTool_mastoidPath}
\end{figure}

This second trial assumes the same conditions as the previous one. It involves the same curved tool and the desired path. However, this trial imposed a unilateral constraint on the tool motion. Consequently, the tool can leave the center point of the incision orifice and move near the orifice wall. This behaviour is demonstrated in Fig.~\ref{fig: exp_pf_ucm_curvedTool_mastoidPath}. The sub-figure (b1) of the latter figure shows the path done by the tool-tip during the outside/transition phases, while the sub-figure (b2) presents the tool-tip path during the inside phase. The dangerous and critical regions are presented by the green and red circles in the latter sub-figure.

\begin{figure}[!h]
	\centering
	\includegraphics[width=.6\columnwidth]{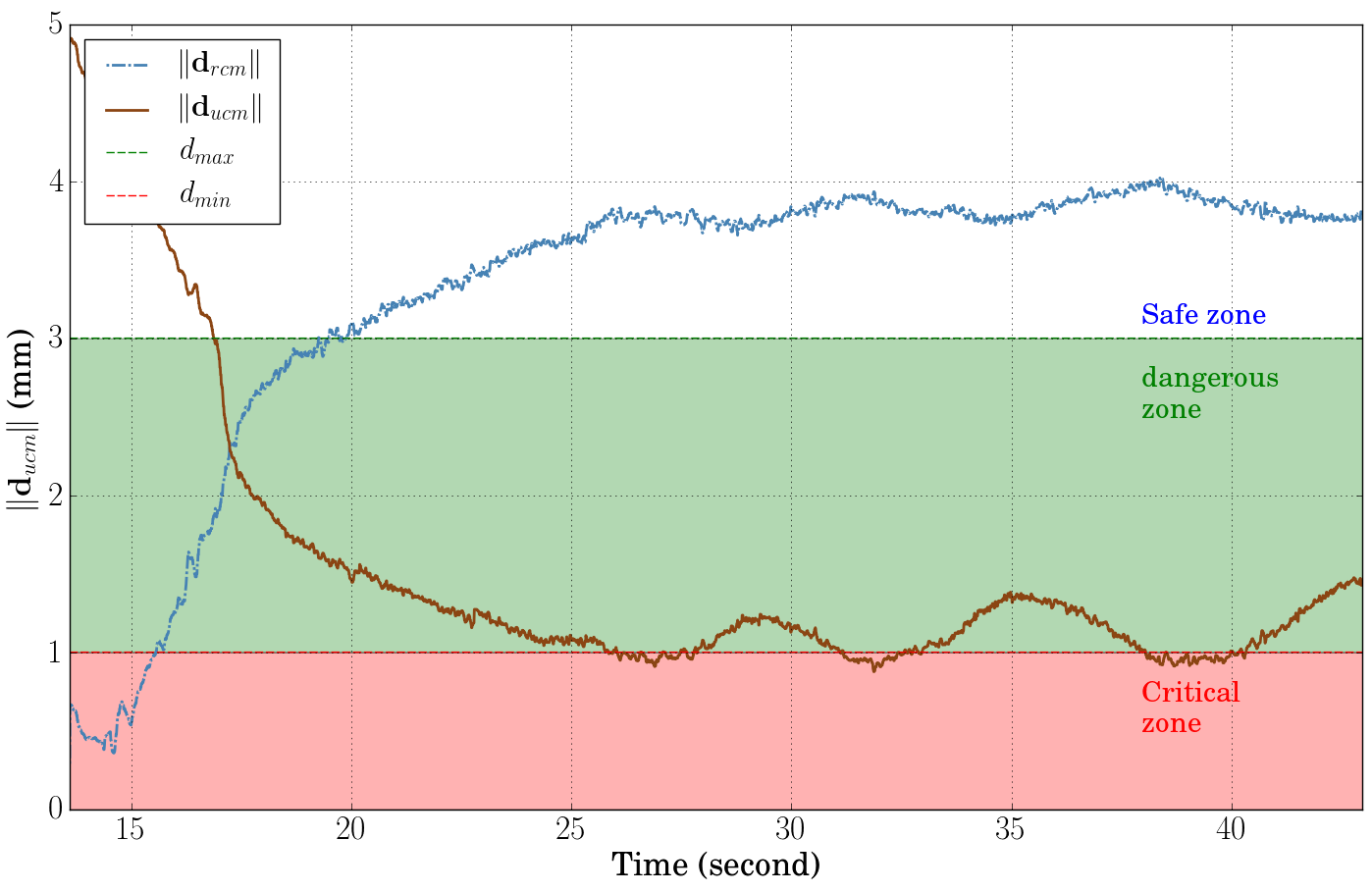}
	\caption{The UCM task error $\mathbf{d}_{ucm}$ during the inside phase.}
	\label{fig: exp_ucm_ct_mp_error__d_ucm}
\end{figure}
\begin{figure}[!h]
	\centering
	\includegraphics[width=.6\columnwidth]{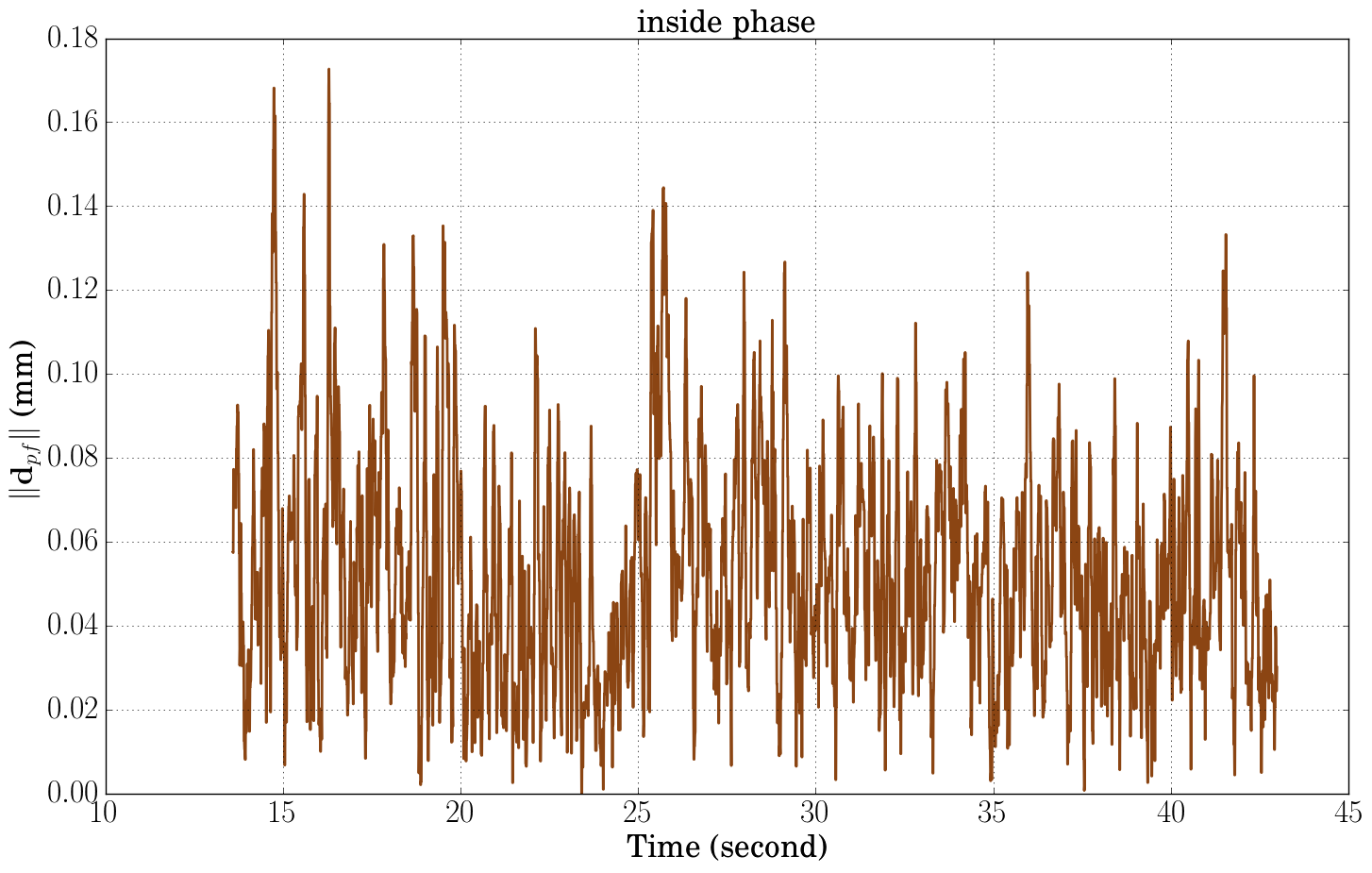}
	\caption{The path-following task error $\mathbf{d}_{pf}$ during the inside phase.}
	\label{fig: exp_ucm_ct_mp_error__d_pf}
\end{figure}

Throughout the inside phase, the hierarchical controller arranges the different tasks as explained in section~\ref{subSec: ucm}. The highest priority is the path-following task when the tool is located in the safe zone. However, the highest priority changes to the UCM task when the tool body passes the danger zone.
The system performances are shown in Fig.~\ref{fig: exp_ucm_ct_mp_error__d_ucm}-\ref{fig: exp_ucm_ct_mp_error__d_pf}. One can observe from the UCM task error $\mathbf{d}_{ucm}$ (Fig.~\ref{fig: exp_ucm_ct_mp_error__d_ucm}) that the tool body is maintained in the dangerous zone since the error $\mathbf{d}_{ucm}$ changes its value between $d_{max}$ and $d_{min}$. 
Besides that, the path-following error $\mathbf{d}_{pf}$ (Fig.~\ref{fig: exp_ucm_ct_mp_error__d_pf}) was $0.05 \pm 0.03 mm$ (mean error $\pm$ STD error) and its median error was $0.05mm$. 

A exteroceptive sensor used as the feedback sensor. Additionally, the gain values used for this second trial were equal to  $\lambda = 1$, $\gamma = 1$, $v_{tis} = 0.5~mm/second$, $\beta^\prime = -1.25$, $\gamma_{c} = -10$ and $T_e = 0.008~second$.

The error $\mathbf{d}_{pf}$ of this trial remains almost the same as the previous trial. It implies that the UCM constraint does not deteriorate the path-following error. Indeed, it provides the surgical tool to move with more liberty in order to take advantage of the large size of the entry orifice.

\begin{table}[!h]
    \centering
    \begin{tabular}{lc c c c }
    \hline
		\rowcolor{sandyBrown}
        N°  & constraint & feedback       & \begin{tabular}[c]{@{}l@{}}type of\\ error\end{tabular}      &  mean ($\left\| \mathbf{e}\right\|$) $\pm$ STD                                                                      \\ \hline\hline
        1  & RCM        & exteroceptive  & \begin{tabular}[c]{@{}l@{}}$d_{rcm}$\\ $d_{pf}$\end{tabular} & \begin{tabular}[c]{@{}l@{}}0.06$\pm$0.05\\ 0.05$\pm$0.02\end{tabular} \\\hline
		\rowcolor{lightSkyBlue}
        2  & RCM        & exteroceptive  & \begin{tabular}[c]{@{}l@{}}$d_{rcm}$\\ $d_{pf}$\end{tabular} & \begin{tabular}[c]{@{}l@{}}0.15$\pm$0.06\\ 0.08$\pm$0.05\end{tabular} \\\hline
        3  & RCM        & proprioceptive & \begin{tabular}[c]{@{}l@{}}$d_{rcm}$\\ $d_{pf}$\end{tabular} & \begin{tabular}[c]{@{}l@{}}0.02$\pm$0.05\\ 0.02$\pm$0.01\end{tabular} \\\hline
		\rowcolor{lightSkyBlue}
        4  & RCM        & proprioceptive & \begin{tabular}[c]{@{}l@{}}$d_{rcm}$\\ $d_{pf}$\end{tabular} & \begin{tabular}[c]{@{}l@{}}0.03$\pm$0.08\\ 0.03$\pm$0.02\end{tabular} \\\hline\hline
        5  & UCM        & exteroceptive  & \begin{tabular}[c]{@{}l@{}}$d_{rcm}$\\ $d_{pf}$\end{tabular} & \begin{tabular}[c]{@{}l@{}}3.30$\pm$0.93\\ 0.05$\pm$0.03\end{tabular} \\\hline
    	\rowcolor{lightSkyBlue}
        6  & UCM        & exteroceptive  & \begin{tabular}[c]{@{}l@{}}$d_{rcm}$\\ $d_{pf}$\end{tabular} & \begin{tabular}[c]{@{}l@{}}3.30$\pm$0.93\\ 0.09$\pm$0.06\end{tabular} \\\hline
        7  & UCM        & proprioceptive & \begin{tabular}[c]{@{}l@{}}$d_{rcm}$\\ $d_{pf}$\end{tabular} & \begin{tabular}[c]{@{}l@{}}2.74$\pm$0.77\\ 0.02$\pm$0.01\end{tabular} \\\hline
		\rowcolor{lightSkyBlue}
        8  & UCM        & proprioceptive & \begin{tabular}[c]{@{}l@{}}$d_{rcm}$\\ $d_{pf}$\end{tabular} & \begin{tabular}[c]{@{}l@{}}2.69$\pm$0.67\\ 0.03$\pm$0.02\end{tabular} \\ \hline\hline
    \end{tabular}
	\caption{Summary of different trials achieved with the curved tool during the experimental tests. \\
    $\left\| \mathbf{e}\right\|$ (in mm) is the absolute average of the linear error along $x-y-z$ axes, and STD is the related standard deviation (in~mm).\\
    Results obtained with the following parameters: $\lambda = 1$, $v_{tis} = 0,5~mm/s$, and $T_e = 0,008~second$. The while trials applied $\beta^\prime = -1.25$, while the blue ones applied $\beta^\prime = -5$.}
	\label{table: expTests}
\end{table}

%
\section{CONCLUSION AND FUTURE WORK}	\label{sec: conclusion}
This article discussed the design of an original controller for guiding a rigid instrument under constrained motions such as RCM or UCM. The proposed methodology allows a generic formulation, in the same controller, two tasks: i) the constrained motion (RCM or UCM), and ii) a revisited 3D path-following scheme by increasing the sensitivity to the path complexity (e.g., curvature radius) and then reducing the path-following error. To manage the achievement of two or more tasks without conflicts, we also implemented a task prioritizing paradigm. Consequently, the developed control scheme can be integrated easily with various robotic systems without an accurate knowledge of the robot inverse kinematics. 

Experimental validation was also successfully conducted using a 6-DoF robotic system. The obtained results are promising in terms of behavior and precision. 
These performances, even if they meet the specifications of the targeted middle ear surgery, may be considered improvements. The positioning error depends directly on the registration process that is not treated optimally in this work. Furthermore, the pose estimation of the tool-tip was done based on a geometric model of the instrument. Its estimation could be another source of error. Thus, it would be interesting to find out another method for estimating the tool shape and the pose of its tip.  

The forthcoming work will implement the discussed methods in a clinical context using a realistic phantom and a human cadaver.
Besides that, a force control could be added to increase the robot sensitivity to its environment and increase the level of security.

\section*{ACKNOWLEDGMENTS}
This work was supported by the Inserm ROBOT Project: ITMO Cancer no~17CP068-00.

\bibliographystyle{plain}
{\small \bibliography{main.bib}}

\end{document}